\documentclass[runningheads]{llncs}
\usepackage{graphicx,epsfig}
\usepackage{amsmath,amssymb} 
\usepackage{multirow}
\usepackage{epsfig}
\usepackage{graphicx}
\usepackage{amsmath}
\usepackage{amssymb}
\usepackage{array}
\usepackage{subcaption,colortbl,color,caption}
\usepackage{tikz}
\usepackage{comment}
\usepackage[pagebackref=true,breaklinks=true,letterpaper=true,colorlinks,bookmarks=false]{hyperref}

\title{Where to Explore Next? ExHistCNN for History-aware Autonomous 3D Exploration} 

\begin{document}
\pagestyle{headings}
\mainmatter
\def\ECCVSubNumber{6579}  

\titlerunning{ExHistCNN for History-aware Autonomous 3D Exploration}
%
\author{Yiming Wang\inst{1}\orcidID{0000-0002-5932-4371} \and
Alessio {Del Bue}\inst{1,2}\orcidID{0000-0002-2262-4872}}
\authorrunning{Y. Wang and A. {Del Bue}}
\institute{Visual Geometry and Modelling (VGM) \and Pattern Analysis and Computer Vision (PAVIS)\\
Fondazione Istituto Italiano di Tecnologia (IIT), Genova, Italy \\
\email{\{yiming.wang, alessio.delbue\}@iit.it}\\
\url{https://github.com/IIT-PAVIS/ExHistCNN}}

\maketitle

\begin{abstract}
In this work we address the problem of autonomous 3D exploration of an unknown indoor environment using a depth camera. We cast the problem as the estimation of the Next Best View (NBV) that maximises the coverage of the unknown area. We do this by re-formulating NBV estimation as a classification problem and we propose a novel learning-based metric that encodes both, the current 3D observation (a depth frame) and the history of the ongoing reconstruction. One of the major contributions of this work is about introducing a new representation for the 3D reconstruction history as an auxiliary utility map which is efficiently coupled with the current depth observation. With both pieces of information, we train a light-weight CNN, named ExHistCNN, that estimates the NBV as a set of directions towards which the depth sensor finds most unexplored areas. We perform extensive evaluation on both synthetic and real room scans demonstrating that the proposed ExHistCNN is able to approach the exploration performance of an oracle using the complete knowledge of the 3D environment.
\end{abstract}

\section{Introduction}
\label{sec:intro}
Being able to perceive the surrounding 3D world is essential for autonomous systems in order to navigate and operate safely in any environment. Often, areas where agents move and interact are unknown, i.e. no previous 3D information is available. There is the need to develop autonomous systems with the ability to explore and cover entirely an environment without human intervention. Even 3D reconstruction from RGBD data is a mature technology \cite{choi2015robust,kinectfusion,henry2012rgb,WhelanSGDL16,meilland2013unifying}, it relies mainly on a user manually moving the camera to cover completely an area. Less attention has been posed to the problem of obtaining a full coverage of the 3D structure of an unknown space without human intervention. This task has strong relations with the longstanding \textit{Next Best View} (NBV) problem~\cite{connolly1985determination,Pito:99,Surmann:etal:2003,Schmid2012} but with the additional issue of having no \textit{a priori} knowledge of the environment where the autonomous system is located. 

\begin{figure*}[!t]
	\begin{center}
		\includegraphics[width=0.8\linewidth]{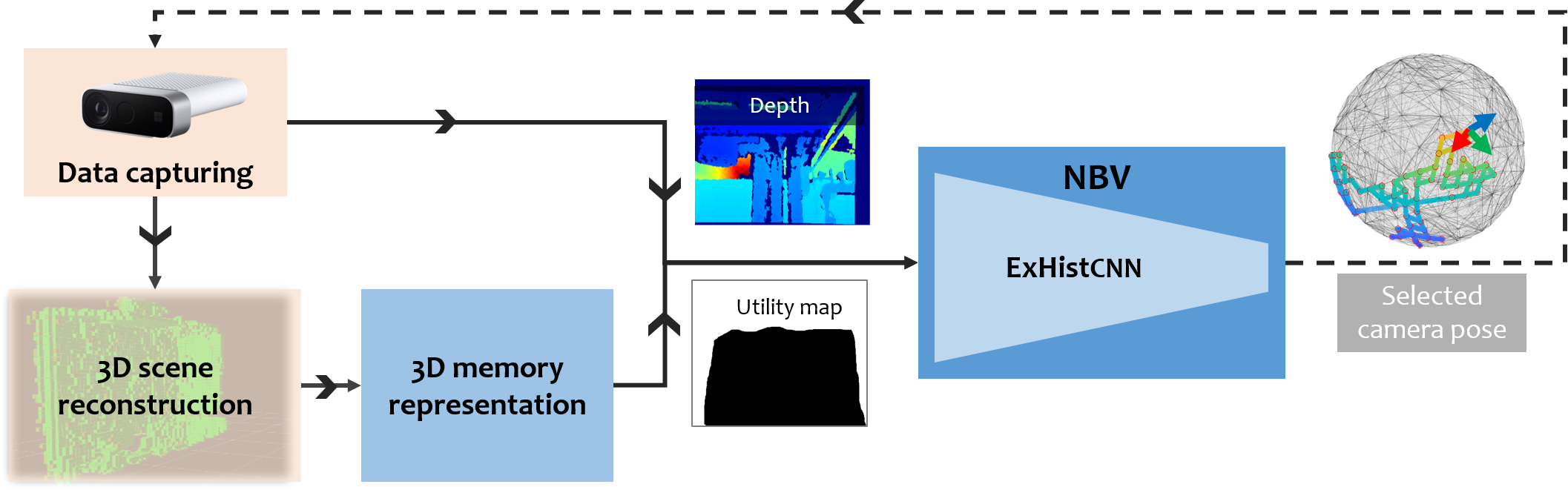}
	\end{center}
	\caption{The overall procedure for autonomous 3D exploration following a Next Best View (NBV) paradigm. Our contributed modules are highlighted in blue.}
	\label{fig:overal_pipeline}
\end{figure*}

In this paper, we address this 3D exploration task using a single depth camera as shown in Fig.~\ref{fig:overal_pipeline}.
At each time step, the system captures a new depth image which is then passed into a general purpose online 3D reconstruction module~\cite{Zhou2018}. The previously reconstructed 3D scene together with the current depth observation provide the NBV module with hints which are often represented as a utility function, whose objective is to select the next view among a set of candidate views that explores most unseen areas of the environment~\cite{Quin2013,potthast2014,Vasquez-Gomez2014,Palazzolo2018,Delmerico2018}. Often the set of candidate views ensure not only the physical reachability but also sufficient overlap with the current view in order to guarantee a feasible 3D reconstruction. How to model the utility is essential in all NBV related literature either in a hand-crafted~\cite{Quin2013,potthast2014,Vasquez-Gomez2014,Palazzolo2018,Delmerico2018} or learning-based manner~\cite{Hepp2018,Wang2019RAL}.  

In such autonomous 3D exploration framework, we propose a novel learning-based NBV method to encode both current observation and reconstruction history with a new CNN model named \textbf{ExHistCNN}. We avoid formulating the task as a regression problem using a 3D CNN \cite{Hepp2018}, since it requires a large number of parameters to optimise along with extensive training data. Instead, our key advantage is that we formulate NBV as a classification problem using a light-weight 2D CNN architecture which needs less training data and computation. 
Our ExHistCNN takes as input the current depth image and the neighbourhood reconstruction status, and outputs the direction that suggests the largest unexplored surface. We exploit ray tracing to produce binary utility maps that encode the neighbourhood reconstruction status and we further propose various data formats to combine the depth and the utility maps to facilitate the history encoding. We train and evaluate our proposed CNN using a novel dataset built on top of the publicly available dataset which are SUNCG~\cite{SUNCG} for synthetic rooms and Matterport3D~\cite{Matterport3D} for real rooms. With experiments, we prove that the proposed CNN and data representation show great potential to encode reconstruction history during 3D exploration and can approach the exploration performance of an oracle strategy with the complete 3D knowledge of the tested environments. The performance is comparable to the state-of-the-art methods~\cite{Quin2013,Wang2019RAL}, with a consistent boost of the scene coverage at the early exploration. 

To summarise, our three major contributions are: 1) We study and evaluate new data embedding to encode history of previously explored areas in the context of NBV estimation; 2) We propose the light-weight ExHistCNN with a careful design of input data, supervision and network, and prove its effectiveness for addressing the 3D exploration problem in unknown environments, and 3) we build a novel dataset based on SUNCG~\cite{SUNCG} and Matterport3D~\cite{Matterport3D} to train and evaluate NBV methods in both synthetic and real environments.

\section{Related work}
\label{sec:soa}
In this section, we will cover related works on NBV (following the observe-decide-move cycle at each step) for 3D exploration and mapping with the focus on the modelling of information utility. 

The information modelling greatly depends on how the 3D environment is represented, which can be categorised as surface-based \cite{borderIcra18} and volume-based representations~\cite{Scott2003}. The volumetric representation is often employed for online motion planning for its compactness and efficiency in visibility operations~\cite{Delmerico2018}. 
Multiple volumetric information metrics have been proposed for selecting the NBV, often through ray tracing. A common idea is to provide statistics on the voxels~\cite{Quin2013,frontier1997,Wettach2010,Vasquez-Gomez2014}, where one can either count the unknown voxels~\cite{Quin2013}, or count only the frontier voxels, which are the voxels on the boundary between the known free space and the unexplored space~\cite{frontier1997,Wettach2010}. Occlusion is further taken into account by counting the \textit{occuplane} (a contraction for occlusion plane) voxels that are defined as bordering free and occluded space~\cite{Vasquez-Gomez2014}. 

In addition to counting-based metrics, there are also metrics based on probabilistic occupancy estimation that accounts for the measurement uncertainty \cite{octomap_2013,Palazzolo2018,mengRAL17}. The main method for computing probabilistic information metrics is based on information entropy~\cite{potthast2014,Kriegel2015,Palazzolo2018,Delmerico2018}. As a ray traverses the map, the information gain of each ray is the accumulated gain of all visible voxels in the form of either a sum~\cite{Delmerico2018} or an average~\cite{Kriegel2015}. The sum favours views with rays traversing deeper into the map, while the average favours more on the uncertainty of voxels regardless the ray depth. Moreover, inaccurate prediction of the new measurement probability during ray tracing can be an issue for computing the information gain if occlusion is not considered. To address this issue, Potthast and Sukhatme~\cite{potthast2014} utilise a Hidden Markov Model to estimate the likelihood of an unknown voxel being visible at any viewpoints. The work in~\cite{Delmerico2018} accounts for the probability of a voxel being occluded via weighting the information gain by the product of the emptiness probability of all voxels before reaching that voxel. Although the computation can be different, the heuristic behind both~\cite{potthast2014,Delmerico2018} is similar, i.e. a voxel with a large unobserved volume between its position and the candidate view position is more likely to be occluded and therefore contributes less information gain.

Recent works have shifted their focus towards learning-based methods \cite{Hepp2018,Wang2019RAL,learn2look2018}. Hepp {\em et al.} train a 3D CNN with the known 3D models of the scene, and the utility is defined as the decrease in uncertainty of surface voxels with a new measurement. The learnt metric shows better mapping performance compared to the hand-crafted information gain metrics. However, the method can be demanding for data preparation and heavy for training. Wang {\em et al.}~\cite{Wang2019RAL} instead propose a 2D CNN to learn the information gain function directly from a single-shot depth image and combine the learnt metric with the hand-crafted metric that encodes the reconstruction history. However due to the fact that a single depth cannot encode reconstruction status, the heuristic-based combination strategies struggle to outperform the strategy using the hand-crafted metric. Jayaraman {\em et al.}~\cite{learn2look2018} address a related but different task, visual observation completion instead of 3D exploration and mapping, where they exploit reinforcement learning trained with RGB images instead of 3D data. 
\section{Proposed Method}
\label{sec:method}
\begin{figure*}[!t]
	\begin{center}
		\includegraphics[width=\linewidth]{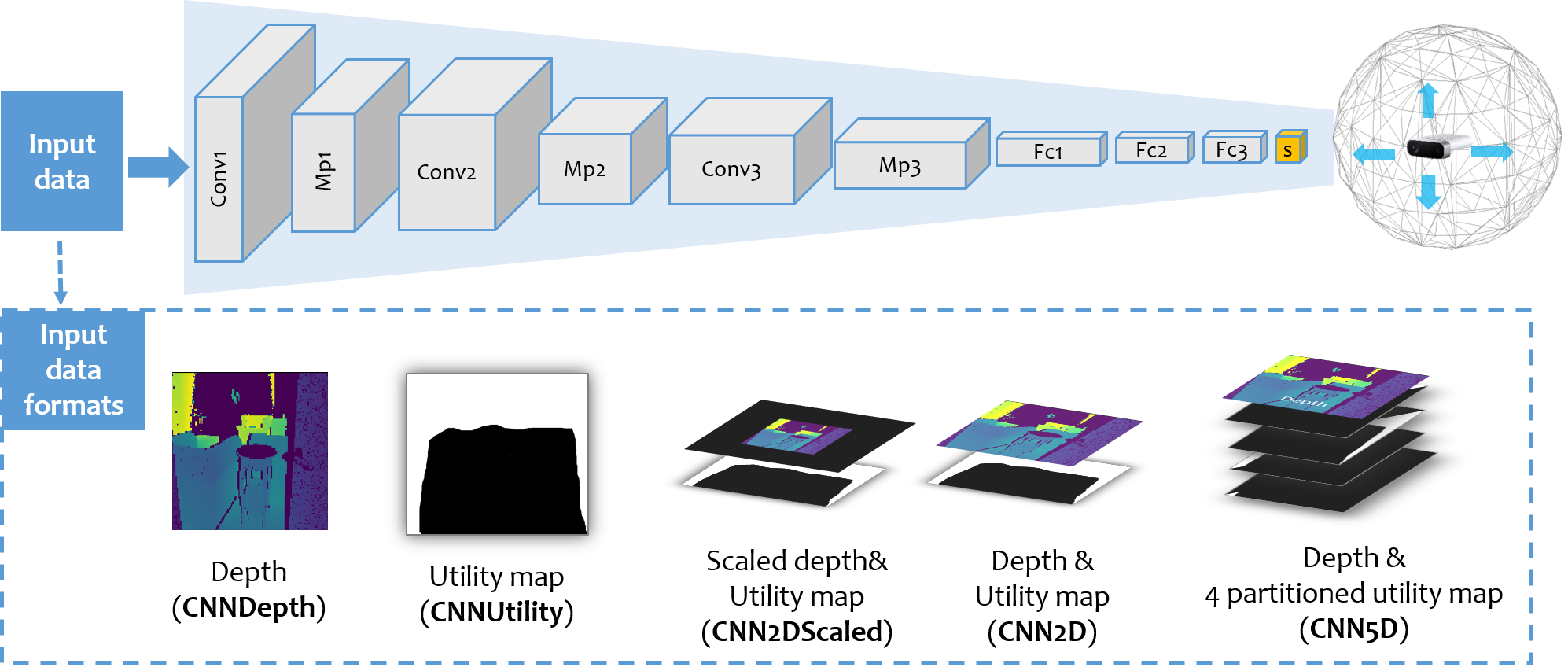}
	\end{center}
	\caption{ExHistCNN architecture and the proposed formats of input data with the aim of encoding local reconstruction status in the context of 3D exploration. 
	}
	\label{fig:cnn_cuboid}
\end{figure*}
We represent the reconstructed scene using octomap~\cite{octomap_2013}, which is an efficient volumetric representation of 3D environments. The space is quantised into voxels that are associated with an occupancy probability $o^i\in\left[0,~1\right]$. A higher $o^i$ value indicates that the voxel is more likely to be occupied, while the lower indicates a higher likelihood to be empty. Based on the occupancy probability, each voxel is thus expressed as having one of the three states: Unknown, Free and Occupied.

At each time step $k$, $o_k^i$ is updated with the new range measurement coming from the depth image $\textbf{D}_{k}$. Let $\textbf{P}_k = \left[\textbf{t}_{k},\textbf{R}_{k}\right]$ be the camera pose at $k$ with $\textbf{t}_{k}$ for translation and $\textbf{R}_{k}$ for rotation. Let $\Omega_k$ be the set of candidate poses that are accessible and also satisfy the view overlapping constraints for 3D reconstruction at $k+1$. The proposed ExHistCNN predicts $m_{k}$, the direction of the NBV which leads to the largest reconstructed surface voxels (see Fig.~\ref{fig:cnn_cuboid}). With the CNN-predicted direction $m_{k}$, the next best pose $\textbf{P}_k^{*}\in \Omega_k$ is selected as the furthest position at direction $m_{k}$ among the candidate poses. 
More specifically, let $\textbf{e}^{m}_{k} = \textbf{R}_{k}\textbf{e}^{m}$ be the unit vector of the $m_{k}$ at the current pose $\textbf{P}_{k}$, where $\textbf{e}^{m}$ is the unit vector of selected direction $m_{k}$ in world coordinate. For each $\textbf{P}_{j}\in \Omega_k$, we define $\Delta\textbf{t}_{j,k} = \textbf{t}_{j} - \textbf{t}_{k}$ as the vector originating from the position of current pose $\textbf{t}_{k}$ and the position of candidate pose $\textbf{t}_{j}$. The projection of $\textbf{e}^{m}_{k}$ and $\Delta\textbf{t}_{j,k}$ can be computed as the dot product $s_{j,k} = \Delta\textbf{t}_{j,k}\textbf{e}^{m}_{k}$. Finally, the pose with the largest projection $s_{j,k}$ is selected as the NBV pose $\textbf{P}_k^{*}$. 

\subsection{Representation for 3D reconstruction history}
\label{sec:method:status}
\begin{figure*}[t!]
\begin{center}
	\begin{tabular}{@{}c}
		\includegraphics[width=0.8\textwidth]{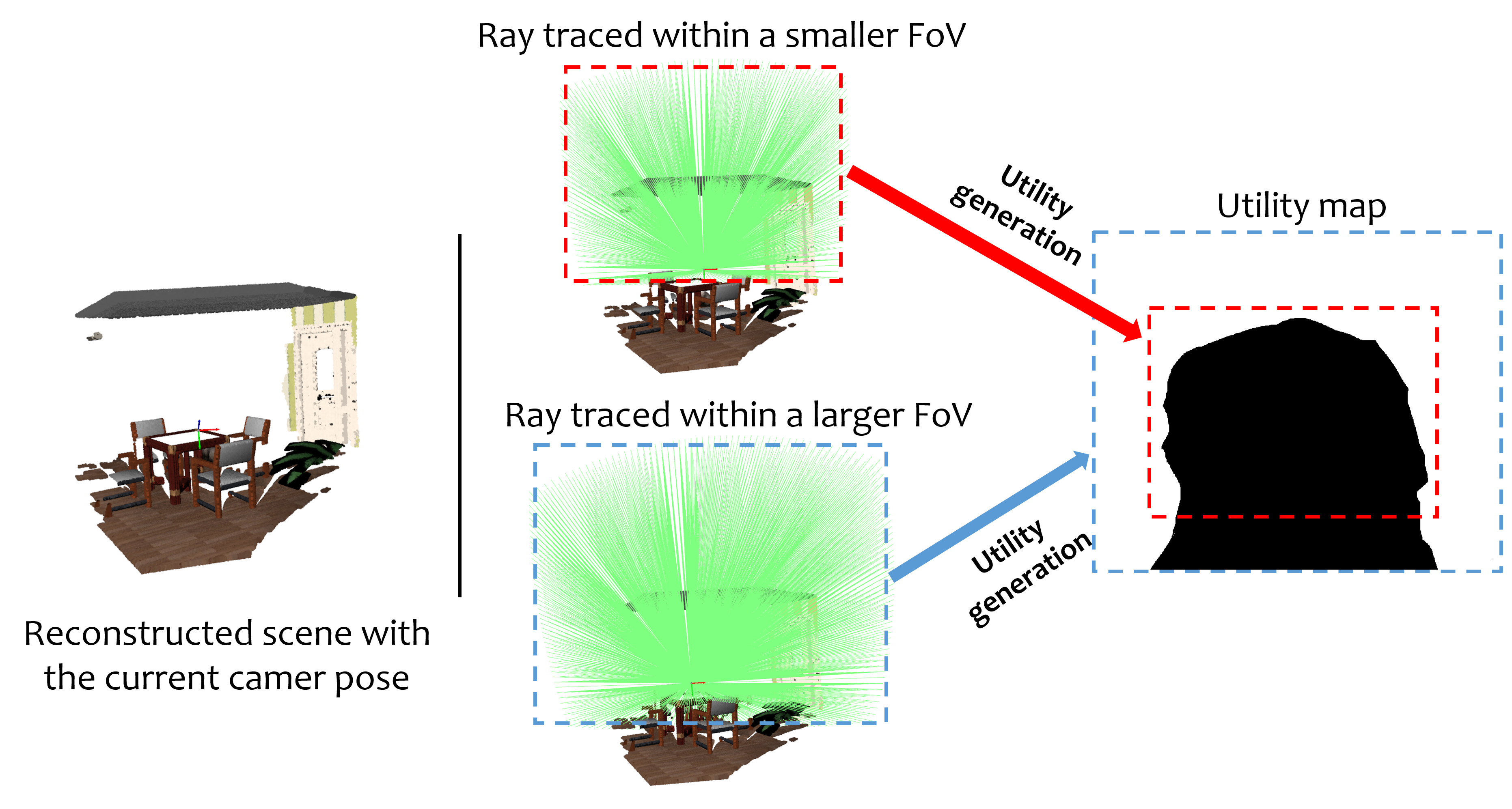}
   	\end{tabular}
\end{center}
\caption{The generation of the binary utility map through ray tracing with different FoV settings. A smaller FoV (highlighted in red) may lead to local solutions when multiple directions expose similar unexplored areas.}
\label{fig:two_step_view}
\end{figure*}
We encode the reconstruction history into a utility map through ray-tracing (see Fig.~\ref{fig:two_step_view}). Given a camera pose $\textbf{P}_k$, we trace a set of rays in a discretised manner within a defined FoV originating from the camera pose towards the 3D space. Each ray corresponds to a utility value. The utility should encourage the NBV towards most unexplored area, we therefore set the utility value to be 1 if the ray does not encounter any Free or Occupied voxels, otherwise 0. The resulting utility map is in practice a binary image and each pixel corresponds to a ray with the value of zero for visited areas, and the value of one for unexplored areas. 

The FoV of ray tracing defines the extension of the reconstruction area that can be used for making the NBV decision. As shown in Fig.~\ref{fig:two_step_view}, the camera can see the area within its one-step neighbourhood (in red box), two-step neighbourhood (in blue box)  or even larger neighbourhoods. We experimentally prove that by seeing the reconstruction status with two-step neighbourhood can improve the exploration performance compared to one-step neighbourhood while keeping the method cost-effective. Note that the utility map in the following sections refers to the FoV that reflects the reconstruction status in two-step neighbourhood.

\subsection{ExHistCNN}
\label{sec:method:ExHistCNN}
The proposed ExHistCNN makes use of a light-weight CNN architecture with convolutional layers, max pooling layers and fully connected layers (see Fig.~\ref{fig:cnn_cuboid}) that functions as a next direction classifier. We consider four main movement directions, i.e. \textit{up, down, left and right}, as the output of the classifier, which is then used to estimate the next camera pose distributed on a sphere surface. Note that the current movement setting is chosen with simplicity for efficient and repeatable datatset preparation and method evaluation. 

In order to obtain the ground-truth direction label $y$, we introduce an \textit{oracle} classifier with the access to ground-truth depth frames at each possible camera movement. In this way, the oracle can always decide the best move that maximises the coverage to unexplored areas at any given pose. ExHistCNN will then learn to imitate the oracle classifier without the access to ground truth depth frames. The network is trained by minimising the cross-entropy loss over the training set $\{\textbf{X}_1, ..., \textbf{X}_{N}\}$ where $N$ is the number of training samples.

Regarding the input $\textbf{X}$, we explored extensively its potential formats and its impact to facilitate the reconstruction history encoding. We firstly investigate the necessity of combining depth and utility map as input $\textbf{X}$ by training basic models using only depth, \textbf{CNNDepth} (Fig.~\ref{fig:cnn_cuboid} (a)), and only utility map, \textbf{CNNUtility} (Fig.~\ref{fig:cnn_cuboid} (b)). Secondly, we further investigate the impact of various strategies in combining the depth image and utility map. As a straightforward option, we stack the depth image and utility map into a two-channel data. Moreover, considering the property of convolution that exploits the spatial information in the data, we train \textbf{CNN2DScaled} (Fig.~\ref{fig:cnn_cuboid} (c)) using the depth that is scaled (in this case, shrunk) based on the ratio of its FoV and the FoV for capturing the utility map with zero padding for the rest of the image, and \textbf{CNN2D} (Fig.~\ref{fig:cnn_cuboid} (d)) using the depth without the scaling.  

\begin{figure}[!t]
\begin{center}
	\begin{tabular}{@{}c}
		\includegraphics[width=0.9\textwidth]{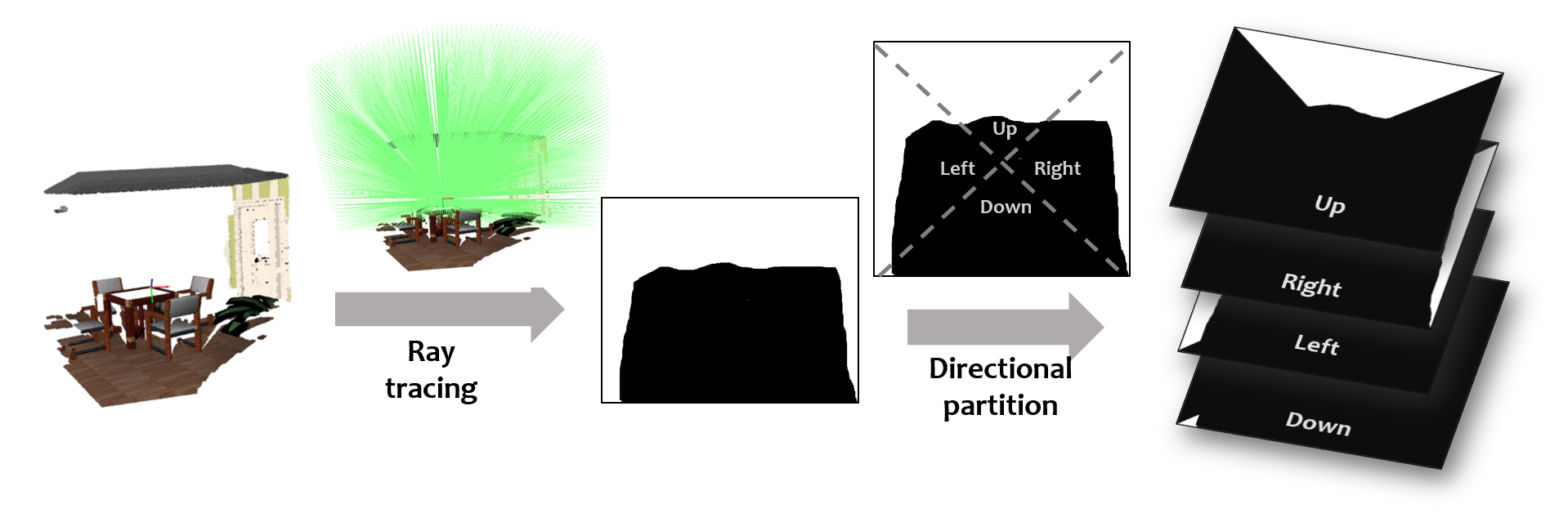}
   	\end{tabular}
\end{center}
\caption{The partitioned utility maps that correspond to the four directions.}
\label{fig:method:division}
\end{figure}

Both \textbf{CNN2D} and \textbf{CNN2DScaled} use the utility map that contains the complete neighbourhood reconstruction history without an explicit indication of where the network should look at. In order to validate if an explicit division of the utility map based on the direction can facilitate the network learning, we also train \textbf{CNN4D} (Fig.~\ref{fig:cnn_cuboid} (e)) and \textbf{CNN5D} (Fig.~\ref{fig:cnn_cuboid} (f)). \textbf{CNN4D} only takes 4 partitioned utility maps with each corresponds to one direction. The partition is performed by dividing the utility map into four non-overlapping triangular areas (see Fig. \ref{fig:method:division}). For each partitioned utility map, the image areas that correspond to other directions will be zero padded in order to not introduce additional information. Such partition choice is experimentally determined because of its better exploration performance compared to another overlapping partition choice. A detailed experiment session for the design choices is provided in Section~\ref{sec:experiments:auto3d}. Finally, \textbf{CNN5D} stacks the depth with the four direction-specific utility maps.

\subsection{NBV for 3D exploration}
\label{sec:method:nbv3d}
In this section, we describe in detail the complete NBV pipeline for 3D exploration of one time step as shown in Fig.~\ref{fig:overal_pipeline}. The system starts with sensor capturing at the time step $k$, where an incoming depth $\textbf{D}_{k}$ (or color-depth pair) frame is then passed into a general purpose online 3D reconstruction module~\cite{Zhou2018} to form a volumetric representation of the scene~\cite{octomap_2013}. Note that we are not bounded to any specific 3D reconstruction algorithm and implementation details are described in Section~\ref{sec:experiments:auto3d}. With the availability of camera poses, which can be obtained by any SLAM algorithm~\cite{WhelanSGDL16}, each depth (or color-depth pair) is registered and integrated to the reconstructed scene volume. We can then obtain the binary utility map by tracing rays into the scene volume with an enlarged FoV in order to have the exploration status for a two-step neighbourhood. We then pass through our proposed ExHistCNN the combined binary utility map and depth frame to predict the best movement direction $m_{k}$ for the next view. 
Given $m_{k}$ predicted by ExHistCNN, the system moves to the selected NBV pose $\textbf{P}_k^{*}$ at the time step $k+1$ and repeats the pipeline, until certain termination criteria are met. In this work we terminate the system once a fixed number of steps is reached to allow most baseline methods to saturate their exploration performance.

\section{Experiments}
\label{sec:experiments}
Section \ref{sec:experiments:cnn:dataset} first describes the dataset generation procedure for the training of ExHistCNN. Then, in Section~\ref{sec:experiments:cnn} we perform the ablation study on the proposed ExHistCNN with various input data formats and network architectures. 
We finally report the 3D exploration performance using both synthetic rooms and real rooms in Section~\ref{sec:experiments:auto3d} with a detailed description on the evaluation dataset and the comparison between our proposed learnt strategies and baseline methods. 

\subsection{Dataset generation}
\label{sec:experiments:cnn:dataset}
We produce a new dataset for training/testing the proposed ExHistCNN using in total 33 rooms from the synthetic SUNCG dataset~\cite{SUNCG}. Our dataset covers various room types including kitchens, living rooms, bedrooms, corridors and toilets. For each room, we rendered a set of viewpoints that are uniformly and isometrically distributed on a sphere with a radius of 20 cm at the height of 1.5m to simulate potential settings of any robotic platforms (e.g. a table-up robotic arm). The camera view is looking out from the sphere centre towards the environment. In particular, we consider a total of 642 viewpoints as the set used for selecting the viewpoints for the tested NBV strategies. 

For each viewpoint, we compute a sub-set of neighbouring viewpoints which are within its circular neighbourhood of a radius $r$. In our experimental setup we set $r$ to 5 cm because the overlapping view constraint is necessary for 3D reconstruction algorithms to work. In order to represent the neighbourhood reconstruction status of each view point in a tractable manner, we discretise status into six levels, i.e. 0\%, 20\%, 40\%, 60\%, 80\% and 100\% of the neighbourhood reconstruction. Each reconstruction level is approximated by selecting the corresponding percentage of neighbouring viewpoints. With the selected view points, we then reconstruct the scene using their corresponding depth frames. For instance, if viewpoint A has 10 neighbouring viewpoints, then 20$\%$ reconstruction status will be achieved by selecting 2 neighbouring viewpoints out of the overall 10 viewpoints for the reconstruction. Since the combinations of the selected viewpoints can be large due to many neighbouring viewpoints, thus to limit the amount of data produced, we constrain only up to 10 combinations for each reconstruction status.

For each viewpoint, we first perform the reconstruction using the neighbouring viewpoints under each neighbouring reconstruction status. We then generate the binary utility map that reflects a two-step neighbourhood reconstruction status through ray tracing. The ground-truth motion label is finally produced by integrating the depth frame that corresponds to the 4 directions in two steps, and selecting the direction that results in most surface voxels. This generation procedure produces 17,960 samples per room, where each sample is composed of a depth image for the current observation, a binary utility map for the neighbourhood reconstruction status and the direction label. We further organise the dataset in a balanced manner with 100K samples per direction class and the train-validation-test follows a 75-15-15 split.

\subsection{ExHistCNN ablation study}
\label{sec:experiments:cnn}
\begin{table}[!t]
\caption{Direction classification result of multiple classifiers at test}
\centering
\resizebox{0.6\textwidth}{!}{
\begin{tabular}{c|c|c|c|c|c|c|c|}
\cline{2-8}
\multirow{2}{*}{} & \multicolumn{4}{c|}{Recall} & \multirow{2}{*}{\begin{tabular}[c]{@{}c@{}}Avg\\ Precision\end{tabular}} & \multirow{2}{*}{\begin{tabular}[c]{@{}c@{}}Avg\\ Recall\end{tabular}} & \multirow{2}{*}{\begin{tabular}[c]{@{}c@{}}Avg\\ F1\end{tabular}} \\ \cline{2-5}
 & up & down & left & right &  &  &\\ \hline
\multicolumn{1}{|c|}{CNNDepth} & 0.469 & 0.58 & 0.32 & 0.41 & 0.446 & 0.445 & 0.445\\ \hline
\multicolumn{1}{|c|}{CNNUtility} & 0.719 & 0.830 & 0.450 & 0.528 & 0.651 & 0.632 & 0.624\\ \hline
\hline
\multicolumn{1}{|c|}{CNN4D} & 0.668 & 0.826 & 0.458 & 0.608 & 0.663 & 0.64 & 0.635\\ \hline
\multicolumn{1}{|c|}{CNN2DScaled} & 0.765 & 0.851 & 0.379 & 0.446 & 0.649 & 0.61 & 0.595\\ \hline
\multicolumn{1}{|c|}{CNN2D} & 0.707 & 0.861 & 0.536 & 0.449 & 0.666 & 0.638 & 0.632\\ \hline
\multicolumn{1}{|c|}{CNN5D} & 0.617 & 0.871 & 0.511 & 0.576 & \textbf{0.677} & 0.644 & 0.642\\ \hline
\hline
\multicolumn{1}{|c|}{MLP4D} & 0.664 & 0.709 & 0.58 & 0.554 & 0.626 & 0.627 & 0.625\\ \hline
\multicolumn{1}{|c|}{MLP2DScaled} & 0.639 & 0.691 & 0.575 & 0.553 & 0.616 & 0.614 & 0.614 \\ \hline
\multicolumn{1}{|c|}{MLP2D} & 0.622 & 0.707 & 0.558 & 0.553 & 0.614 & 0.61 & 0.610\\ \hline
\multicolumn{1}{|c|}{MLP5D} & 0.683 & 0.723 & 0.618 & 0.595 & 0.655 & \textbf{0.654} & \textbf{0.654}\\ \hline
\end{tabular}}
\label{table:classification_accu}
\end{table}
We train ExHistCNN with various input data as described in Section \ref{sec:method:ExHistCNN}, namely, CNNDepth, CNNUtility, CNN2DScaled, CNN2D, CNN4D, and CNN5D. 
As comparison, we also train a set of classifiers using ResNet101~\cite{resnet} pretrained on ImageNet, that serves as a feature extractor. Each channel of the input (see Fig. \ref{fig:cnn_cuboid}) is repeated to 3 channels and fed to ResNet101. The extracted feature vectors are then concatenated and used as input to train a four-layered Multiple Layer Perceptron (MLP) classifier. According to the data input formats, we therefor train four MLP-based classifiers: \textbf{MLP2DScaled}, \textbf{MLP2D}, \textbf{MLP4D} and \textbf{MLP5D}. For all networks, we resize the input to $64\times64$. We apply techniques including batch norm and drop out during training and the batch size is set to maximise the usage of GPU. Stochastic gradient descent is used with learning rate $1e^{-3}$, 200 epochs and momentum 0.9. For testing, we use the model at the epoch where each network starts to saturate. 

Table.~\ref{table:classification_accu} shows the testing classification performance of multiple ExHistCNN models and MLP-based models. Regarding the average classification performance, CNNDepth performs the worst among all. In general, models that combine both depth and utility maps (CNN2D and CNN5D) are better than the models using only utility maps (CNNUtility and CNN4D). Moreover, we notice that partitioning the utility map into four directions, CNN4D is able to perform better than CNNUtility which uses only a single-channel utility map. Similarly, CNN5D with depth and partitioned utility maps is also marginally better than CNN2D. Interestingly, we observe that CNN2DScaled is not performing better than CNN2D. The reason might be that the depth after rescaling and resizing to $64 \times 64$, becomes a rather small patch which could be not very informative for the network to learn from. Moreover, we do notice that all the models perform better in particular directions, i.e. up and down is better than left and right. This can be due to the standard camera setting with wider horizontal FoV than the vertical FoV. Finally, the MLP-based models have a similar pattern as our CNN models, however are achieving a worse classification performances apart from MLP5D. One possible reason can be that the pretrained network extracts feature vectors with semantics bias from other datasets, while our CNN models is trained from scratch without the impact of external bias. In the following section we will use CNN2D and CNN5D as our ExHistCNN models for their best performance, to evaluate the 3D exploration performance. Other models including CNNUtility, CNN4D, MLP4D, MLP2D and MLP5D are also evaluated. 

\subsection{Autonomous 3D exploration performance}
\label{sec:experiments:auto3d}
We apply different NBV strategies to indoor dataset for 3D exploration and report the surface coverage ratio, i.e. the number of the surface voxels generated using autonomous methods against the number of surface voxels of a complete reconstructed room. For a fair comparison, we evaluate  all methods until a fixed number of steps (150 steps throughout the experiments). The step number is set to allow most strategies to saturate in their exploration performance, i.e. when the camera starts looping within a small area. The metric that achieves a larger coverage ratio within the fixed number of steps is considered better.
\begin{figure}[t!]
\begin{center}
	\begin{tabular}{@{}c@{}p{0.05\textwidth}@{}c}
		\includegraphics[width=0.45\textwidth]{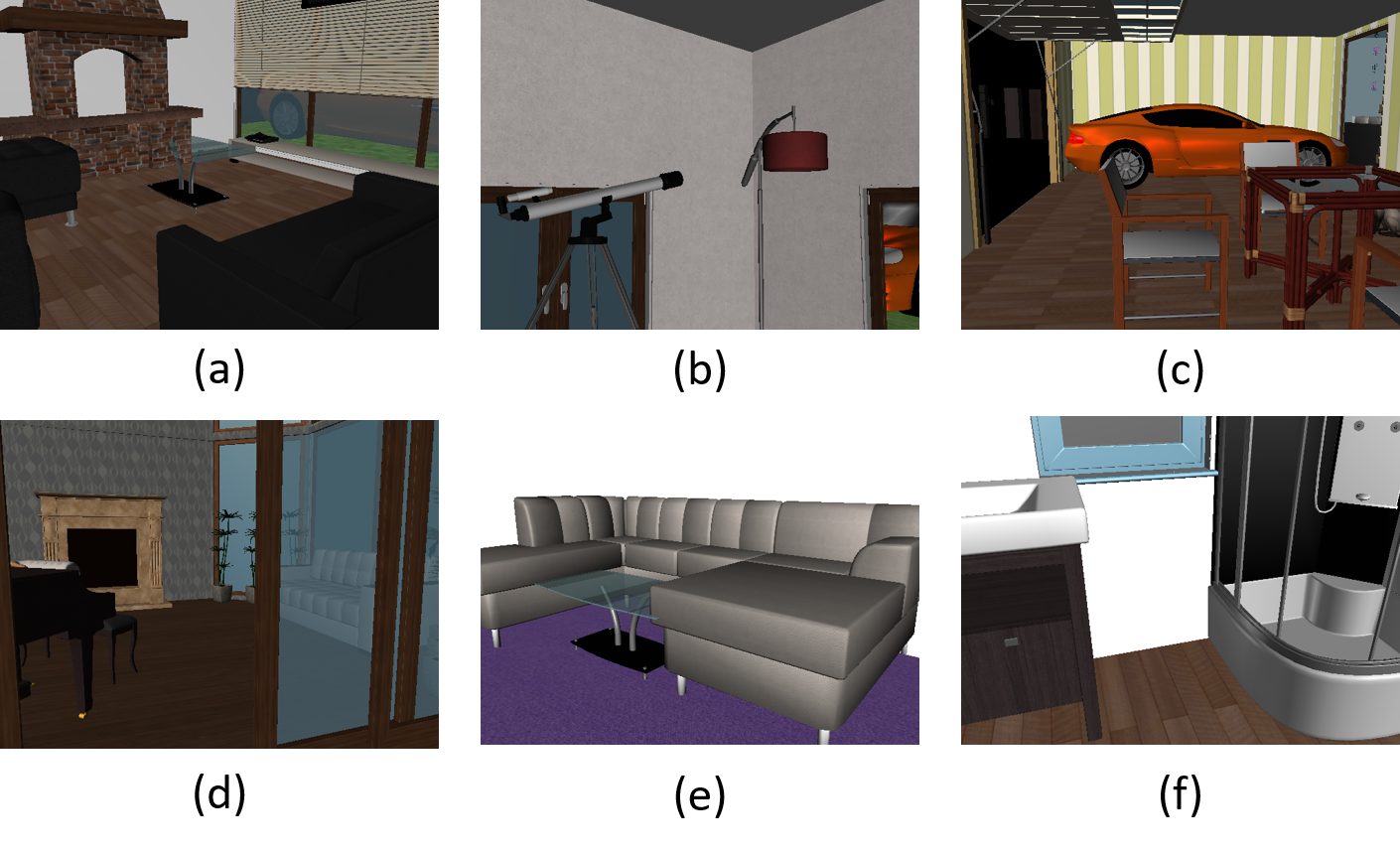}& &\includegraphics[width=0.45\textwidth]{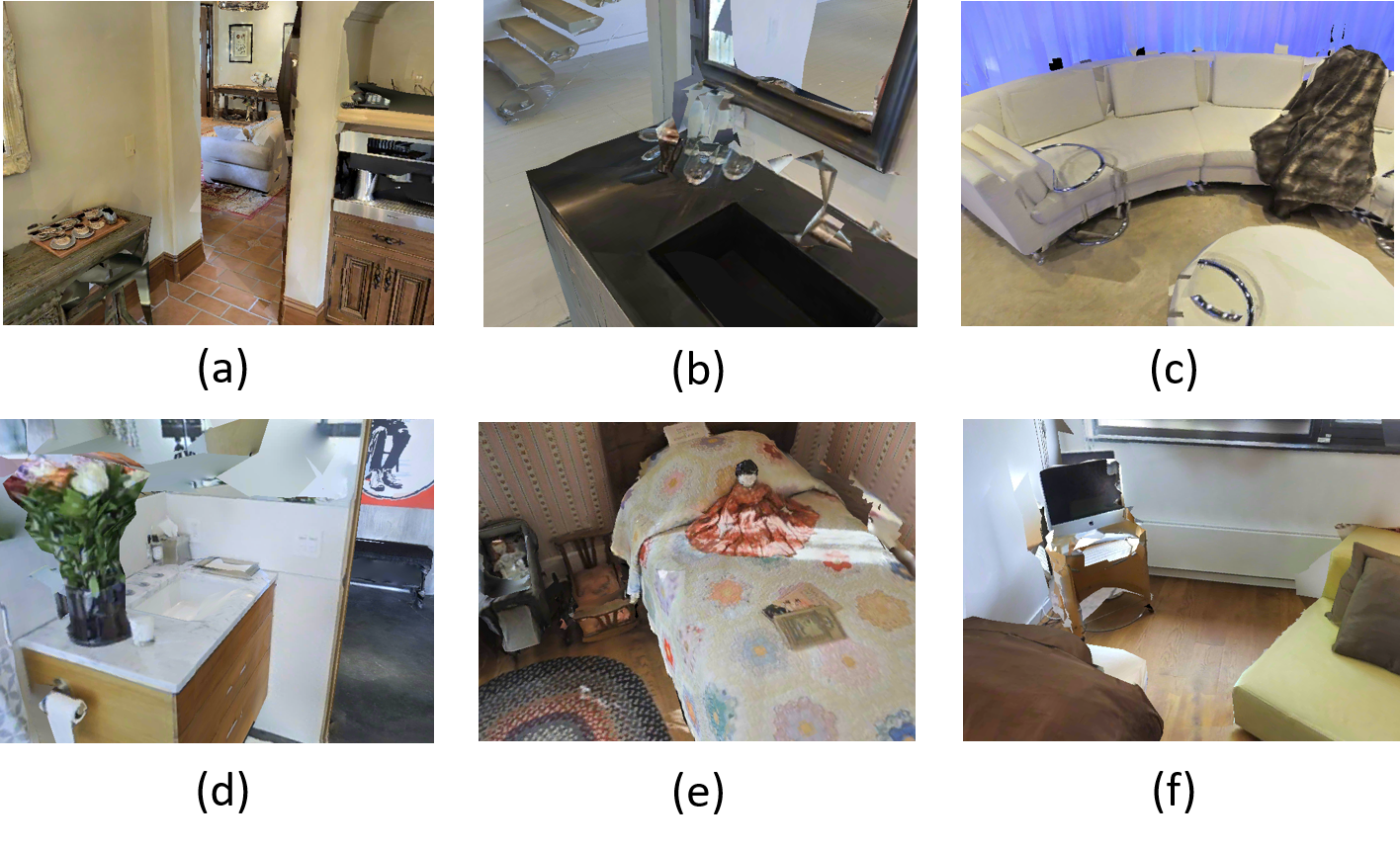}\\
		(a) SUNCG synthetic scenes & & (b) Matterport3D real scene scans
   	\end{tabular}
\end{center}
\caption{Rooms used for the exploration experiment. (a) six synthetic rooms from SUNCG dataset rendered with SUNCG Toolbox.  (b) six real room scans from Matterport3D dataset rendered with HabitatSim.}
\label{fig:exp:syn_setup}
\end{figure}

\paragraph{\textbf{Evaluation dataset.}}We perform experiments using dataset rendered from synthetic rooms in SUNCG dataset~\cite{SUNCG} (Fig.~\ref{fig:exp:syn_setup} (a)) and real room scans (Fig.~\ref{fig:exp:syn_setup} (b)) from Matterport3D~\cite{Matterport3D} using HabitatSim~\cite{HabitatSim}. The set of synthetic rooms for evaluating the exploration performance is different than the set of rooms used for training our ExHistCNN models. Moreover, to validate the generalisation of the model from synthetic to more realistic data, 
we use a publicly available tool, HabitatSim, to render depth (and color) data with real room scans from Matterposrt3D, following the same dome-shaped path as described in Section~\ref{sec:experiments:cnn:dataset}. 
In the experiments, scenes are reconstructed using a truncated signed distance function (TSDF) volume integration method with implementation tools provided in Open3D~\cite{Zhou2018}. 
In particular, we consider that the dome-shaped path explores the complete room, i.e. the coverage ratio is $100\%$.

\begin{figure}[!t]
\begin{center}
	\begin{tabular}{@{}c}
		\includegraphics[width=0.6\textwidth]{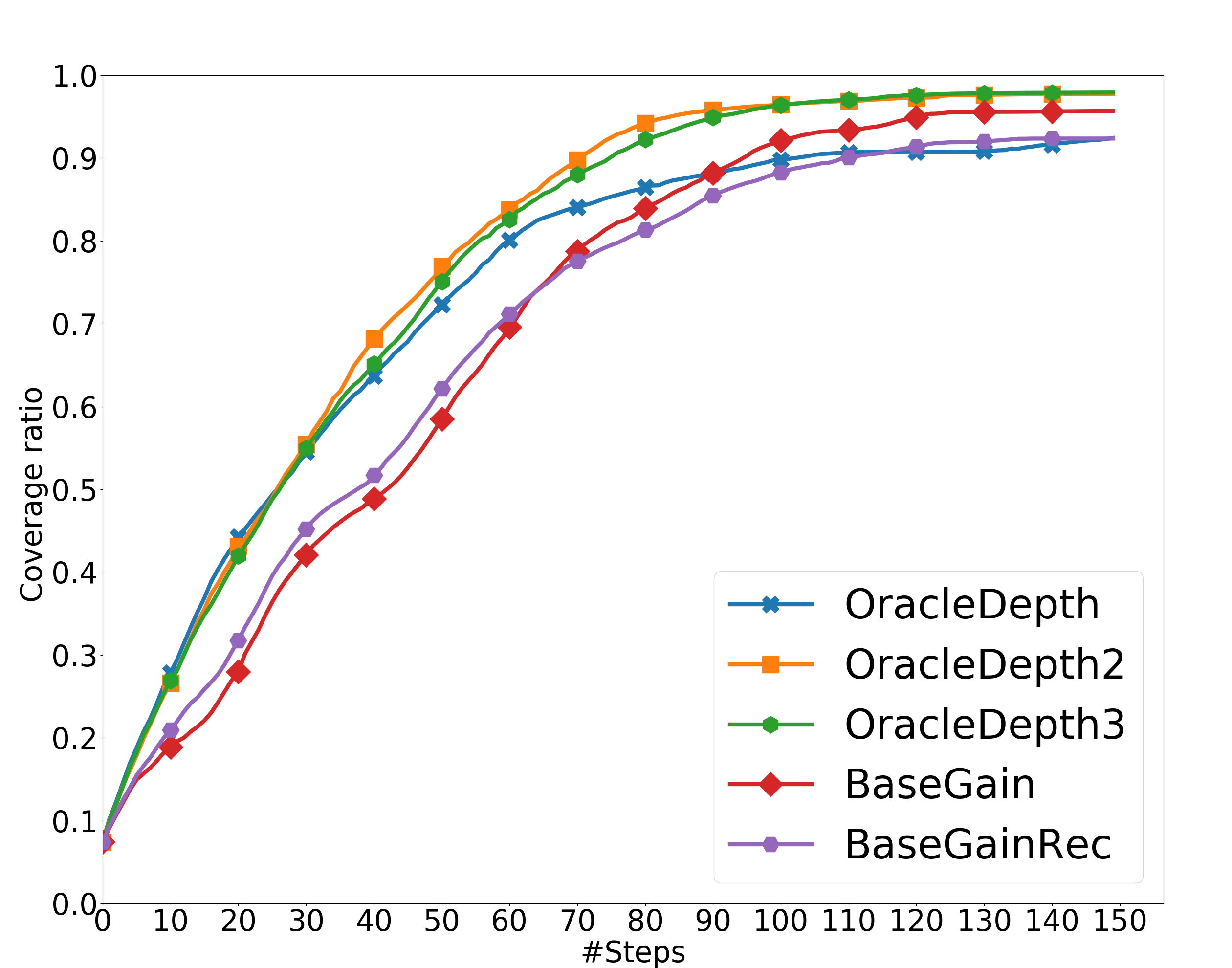}
   	\end{tabular}
\end{center}
\caption{The coverage ratio with time achieved with strategies that are used for justifying design choices.}
\label{fig:experiments:auto3d:ablation}
\end{figure}
\paragraph{\textbf{Justification of design choices.}}We first performed a set of experiments to justify two choices in the method and dataset design: 1) the FoV selection during ray-tracing to produce the utility map that reflects multiple-step neighbourhood reconstruction status and 2) the partitioning of the utility map to enforce the directions in the input data. To justify the FoV selection, we perform 3D exploration using the oracle NBV strategies by integrating the depth frames in different time steps. \textbf{OracleDepth} integrates the depth frame corresponding to each candidate direction for the next step into the current volume, and the NBV is selected with the largest resulted surface voxels. Similarly, \textbf{OracleDepth2} (\textbf{OracleDepth3}) integrates the depth frames corresponding to each candidate direction for the next two (three) steps into the current volume. To justify the partitioning of the utility map, we perform 3D exploration using the NBV selected based on the sum of each partitioned utility map. \textbf{BaseGain} divides the utility map into four non-overlapping triangular areas that correspond to the four candidate directions (see Fig.~\ref{fig:method:division}), while \textbf{BaseGainRec} divides the utility map by half for each direction, resulting in rectangular overlapping areas.

Fig.~\ref{fig:experiments:auto3d:ablation} shows the coverage ratio with time using the above motioned strategies. Results are averaged by five independent runs on the six synthetic rooms. We observe that OracleDepth2 and OracleDepth3 outperform OracleDepth because being aware only one-step ahead the reconstruction status can be prone to early saturation due to local solutions. OracleDepth2 achieves almost the same exploration speed and coverage performance as OracleDepth3, but with reduced computational/storage cost for both offline dataset preparation and NBV estimation at runtime. We therefore perform ray-tracing to produce the utility map that reflects the two-step neighbourhood reconstruction status. Moreover, BaseGain achieves a higher coverage ratio compared to BaseGainRec, which makes the non-overlapping triangular partition a better choice.

\begin{figure}[t!]
\begin{center}
	\begin{tabular}{@{}c@{}c}
		\includegraphics[width=0.48\textwidth]{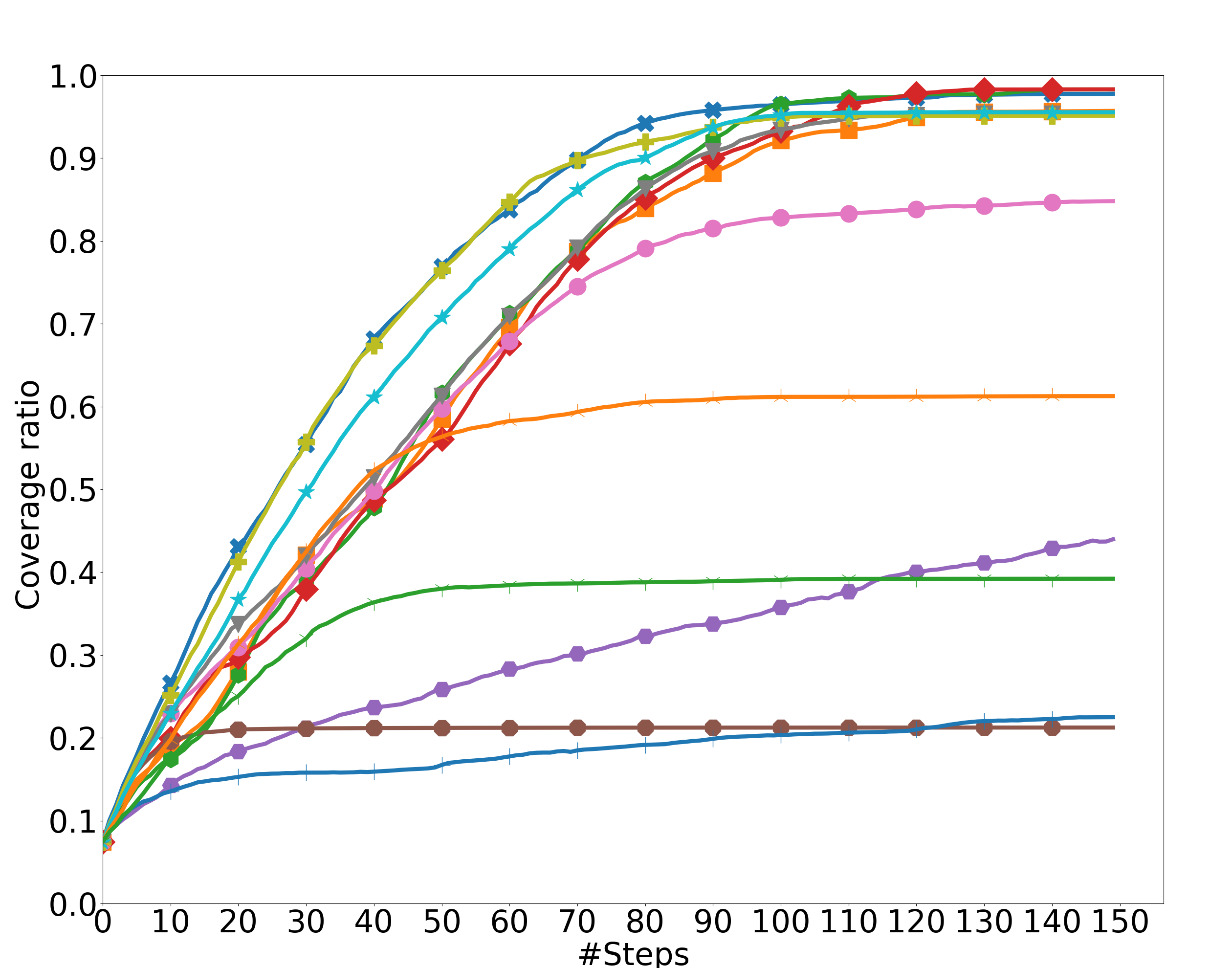}&\includegraphics[width=0.48\textwidth]{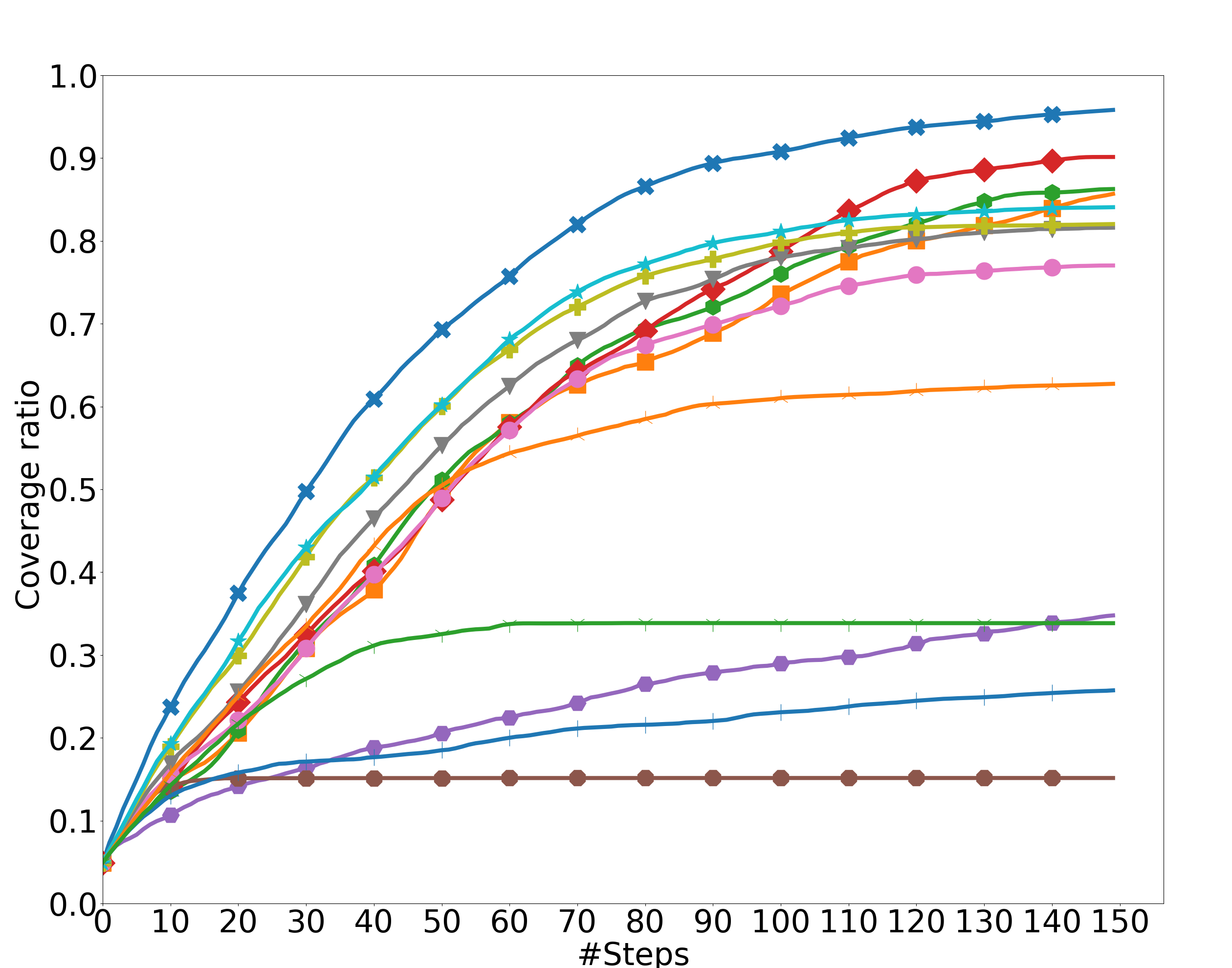}\\
		(a) synthetic SUNCG rooms &(b) Matterport3D real room scans\\
		\includegraphics[width=0.48\textwidth]{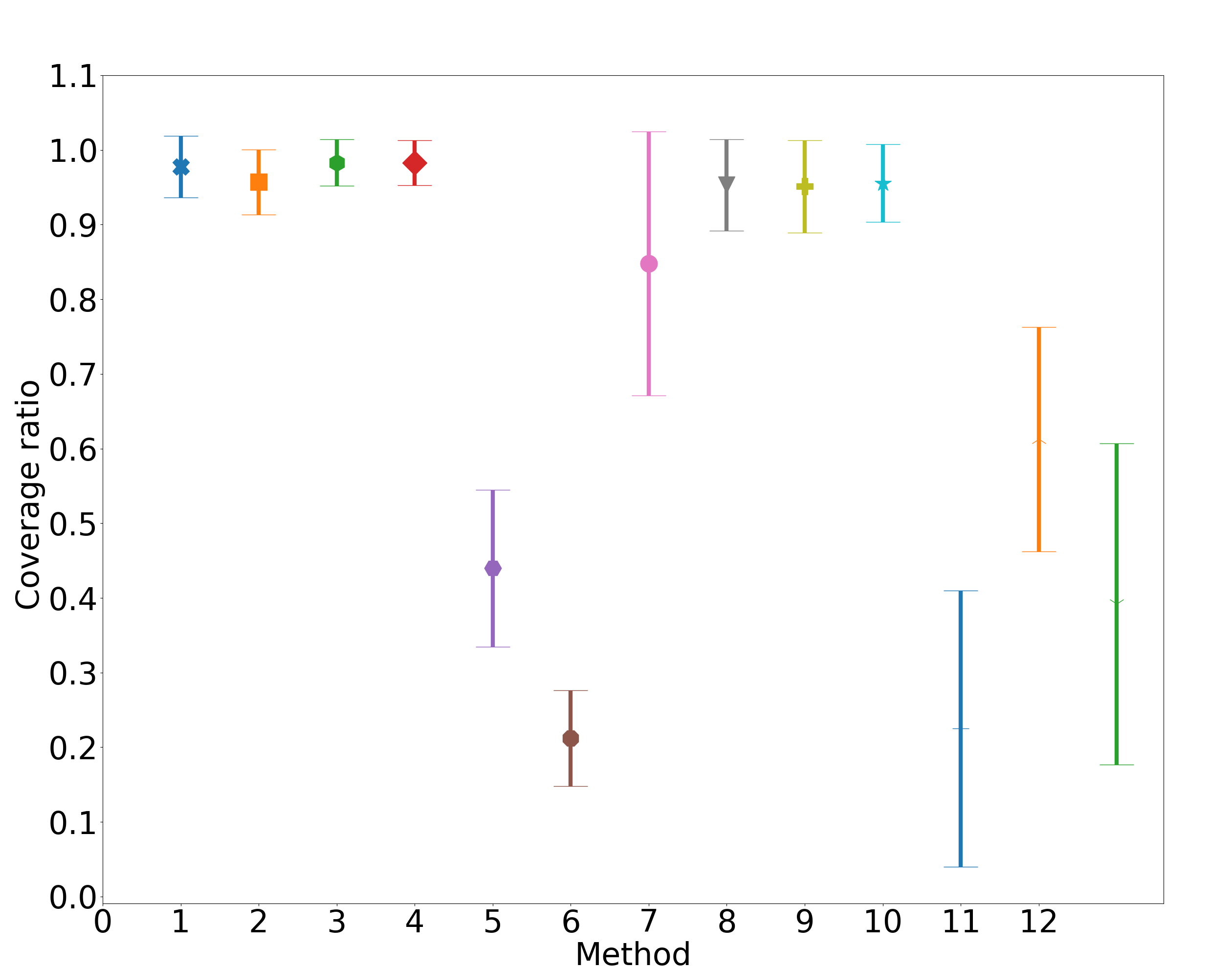}&\includegraphics[width=0.48\textwidth]{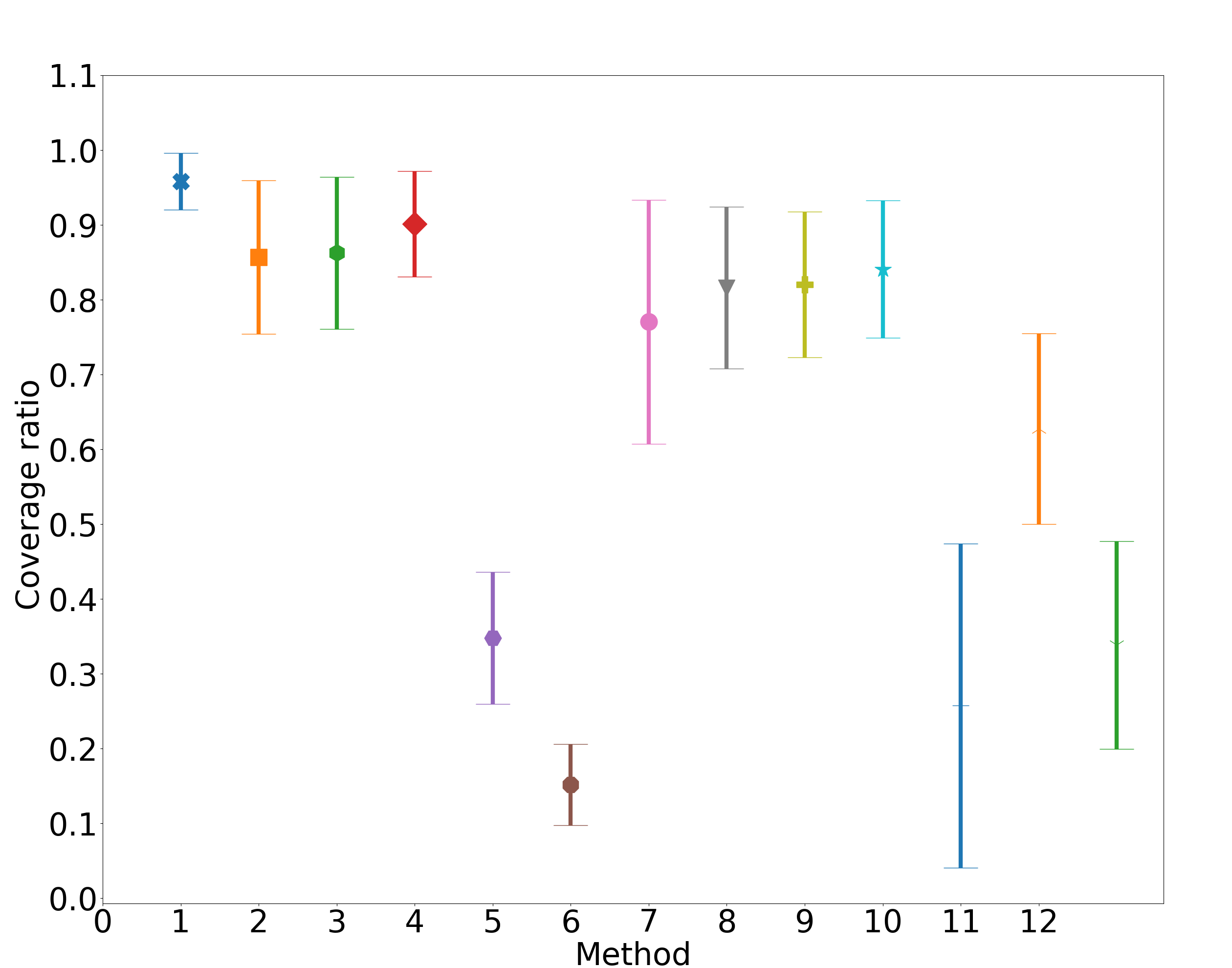}\\
		\multicolumn{2}{c}{\includegraphics[width=0.8\textwidth]{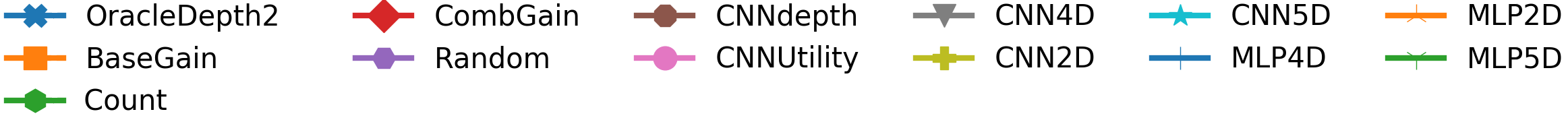}} \\
		(c) synthetic SUNCG rooms &(d) Matterport3D real room scans
   	\end{tabular}
\end{center}
\caption{The coverage ratio with time (upper row) and the averaged final coverage ratio and its standard deviation (lower row).}
\label{fig:exp:coverage_ratio_syn}
\end{figure}

\paragraph{\textbf{Methods and baselines comparison.}}We performed the exploration experiments using our ExHistCNN models with various input: \textbf{CNNdepth}, \textbf{CNNUtility}, \textbf{CNN4D}, \textbf{CNN2D} and \textbf{CNN5D}, as well as the MLP-based models: \textbf{MLP4D}, \textbf{MLP2D} and \textbf{MLP5D} as described in Section~\ref{sec:experiments:cnn}. We compared the above mentioned learning-based strategies against: \textbf{Random} strategy that randomly selects the NBV, \textbf{BaseGain} that selects the view based on the sum of the partitioned utility maps, \textbf{Count} \cite{Quin2013} that selects the NBV by counting unknown voxels for each candidate pose, and \textbf{CombGain}~\cite{Wang2019RAL} that selects the NBV using both the output of CNNDepth for direction and the entropy-based utility maps computed using view-dependent descriptors for each candidate pose\footnote{We are not able to compare with \cite{Hepp2018} as their dataset and code are not available.}. Finally, \textbf{OracleDepth2} serves as a reference for the best reachable result for our learning-based strategies.

Fig.~\ref{fig:exp:coverage_ratio_syn}(a) shows the average coverage ratio over time for synthetic rooms. NBV with CNNDepth achieves the worst coverage because the camera moves without any knowledge of the reconstructions status, leading to repeated back and forth movement at a very early stage. We observe that CNNUtility with only the binary utility map as its input is worse than CNN4D which uses the partitioned utility maps. This result indicates that the partition of input data can facilitate the network learning of the NBV directions and boost the exploration performance. Moreover, our variants CNN2D and CNN5D leverage the depth information to explore faster compared to CNN4D at the earlier stage, and saturate at a similar coverage ratio which approaches to the performance of OracleDepth2. In the early phase, CNN2D explores faster than CNN5D at a similar exploration speed as OracleDepth2. MLP-based strategies in general are worse than the ExHistCNN-based strategies. Even the best performed MLP2D model is almost 30\% less than CNN2D. BaseGain, Count and CombGain use hand-crafted utility, and are slower in the beginning, however Count and CombGain are able to achieve a slightly better coverage ratio. 
\begin{figure*}[t!]
\begin{center}
	\begin{tabular}{@{}c@{}c@{}c@{}c}
		{\em Random} & {\em CNNDepth}& {\em CNNUtility} & {\em OracleDepth2}\\
		\includegraphics[width=0.20\textwidth]{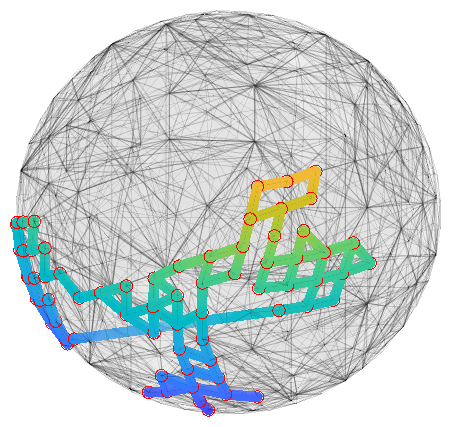}& \includegraphics[width=0.20\textwidth]{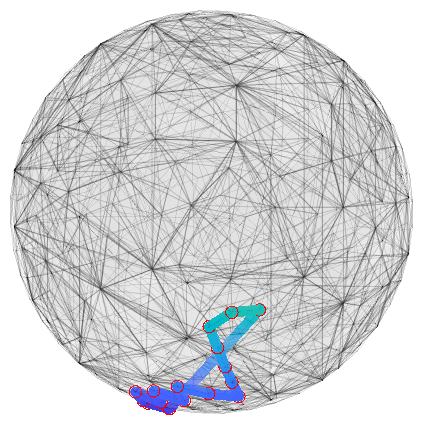}&
		\includegraphics[width=0.20\textwidth]{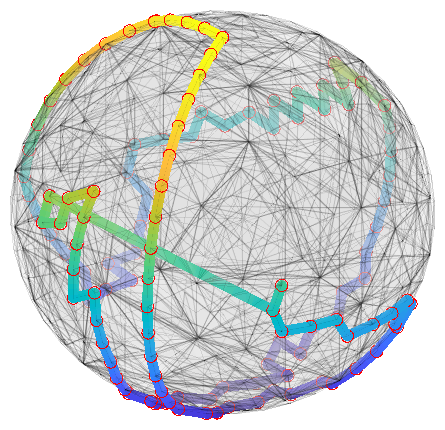}&
		\includegraphics[width=0.20\textwidth]{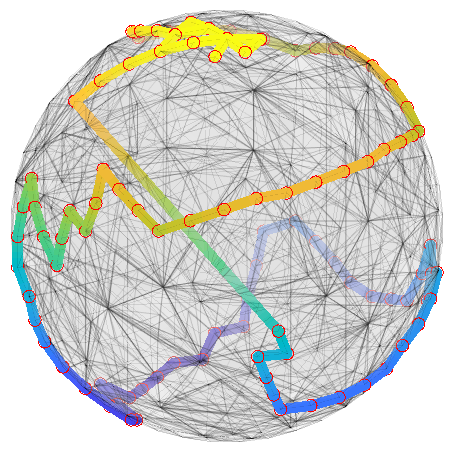}
		\\
		\includegraphics[width=0.20\textwidth]{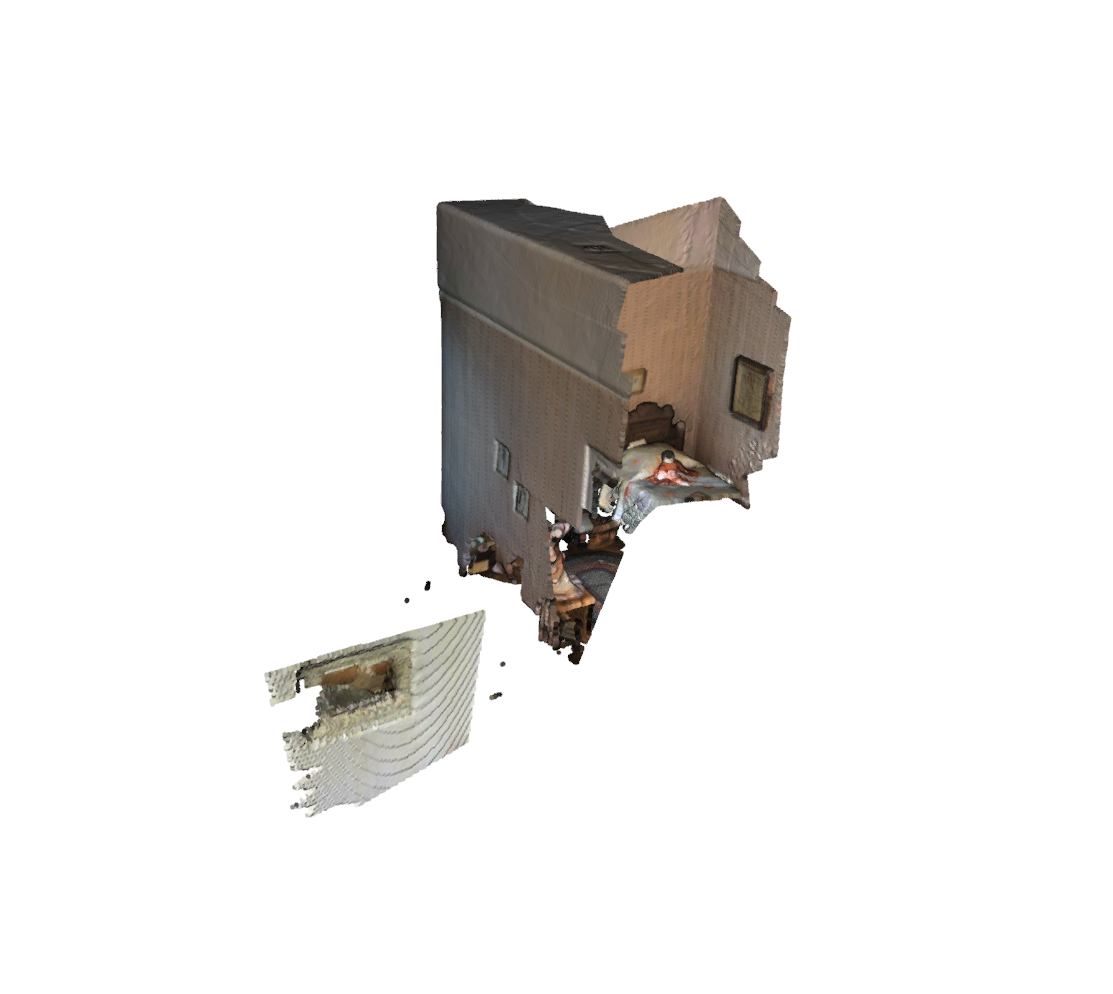} &
		\includegraphics[width=0.20\textwidth]{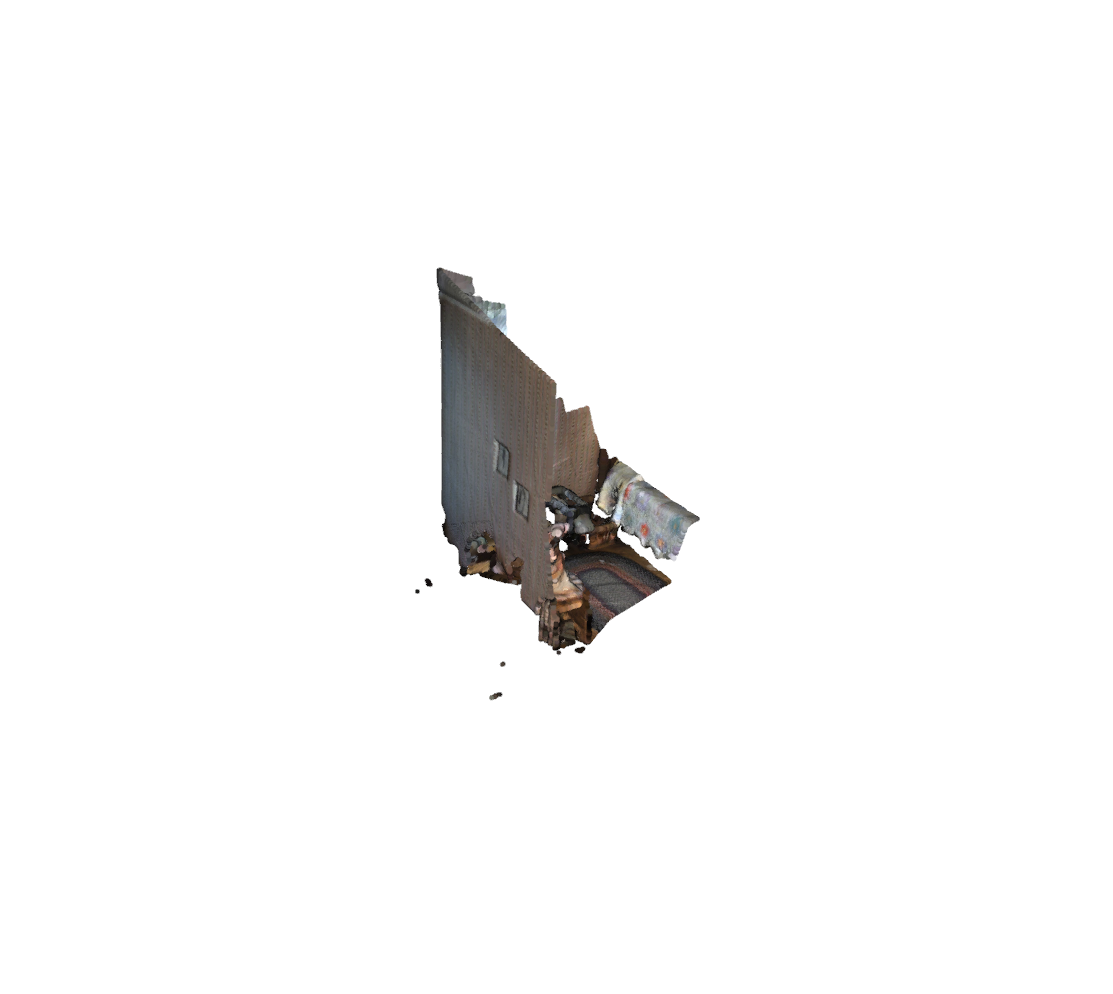}&
		\includegraphics[width=0.20\textwidth]{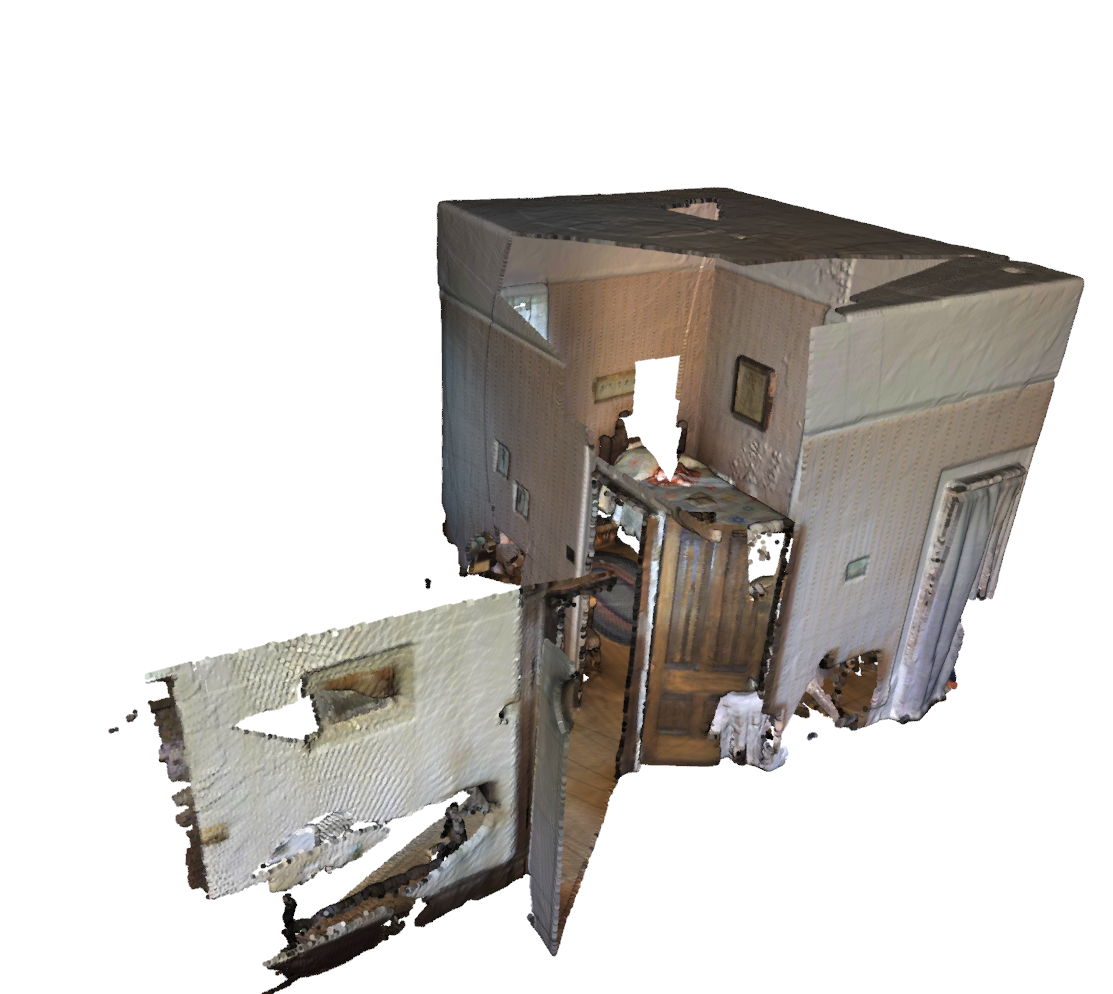}&
		\includegraphics[width=0.20\textwidth]{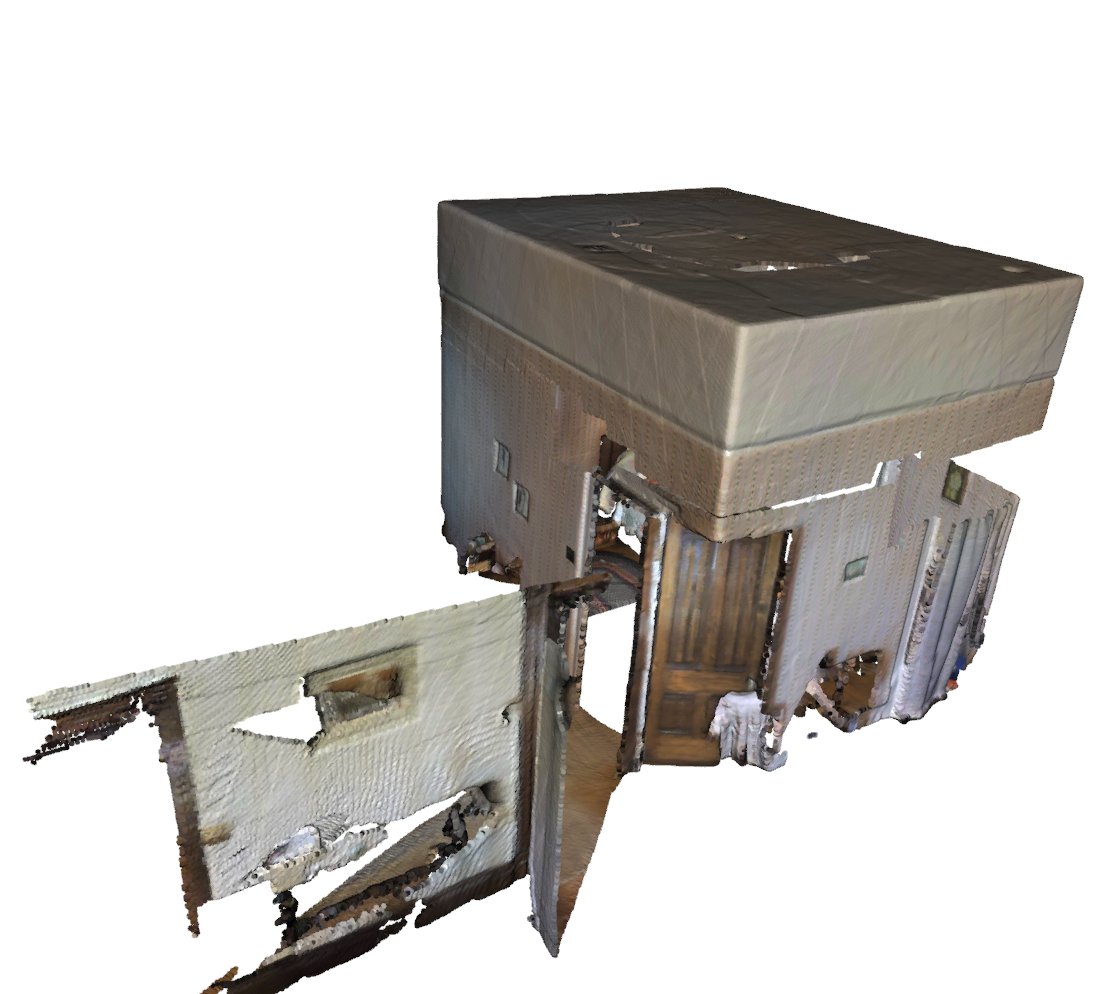}
		\\
		\hline
		\\
		{\em BaseGain} & {\em CombGain} & {\em CNN2D}& {\em CNN5D}	
		\\
		\includegraphics[width=0.20\textwidth]{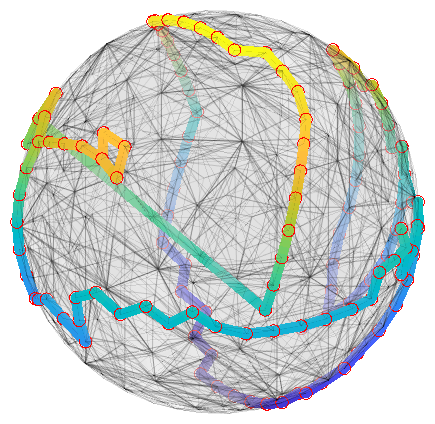}&
	     \includegraphics[width=0.20\textwidth]{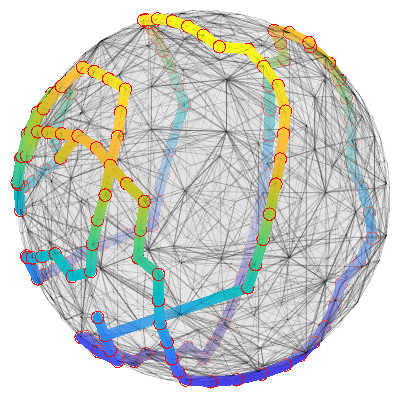}&
		\includegraphics[width=0.20\textwidth]{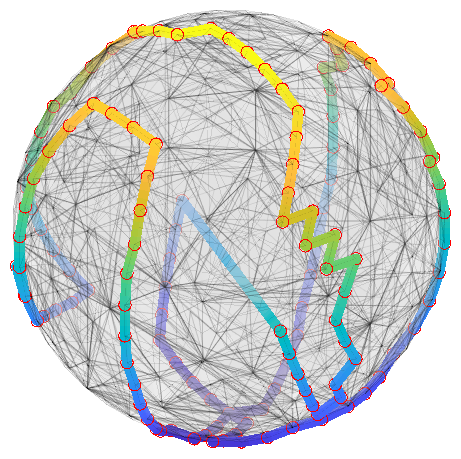}&
		\includegraphics[width=0.20\textwidth]{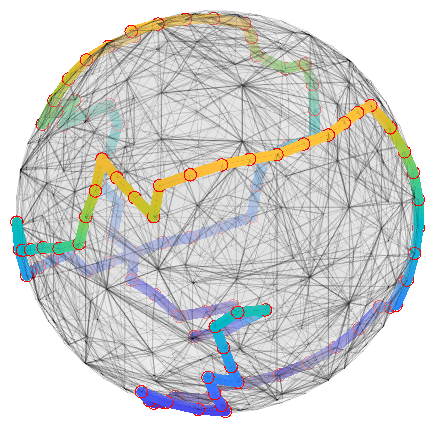}
		\\
		\includegraphics[width=0.20\textwidth]{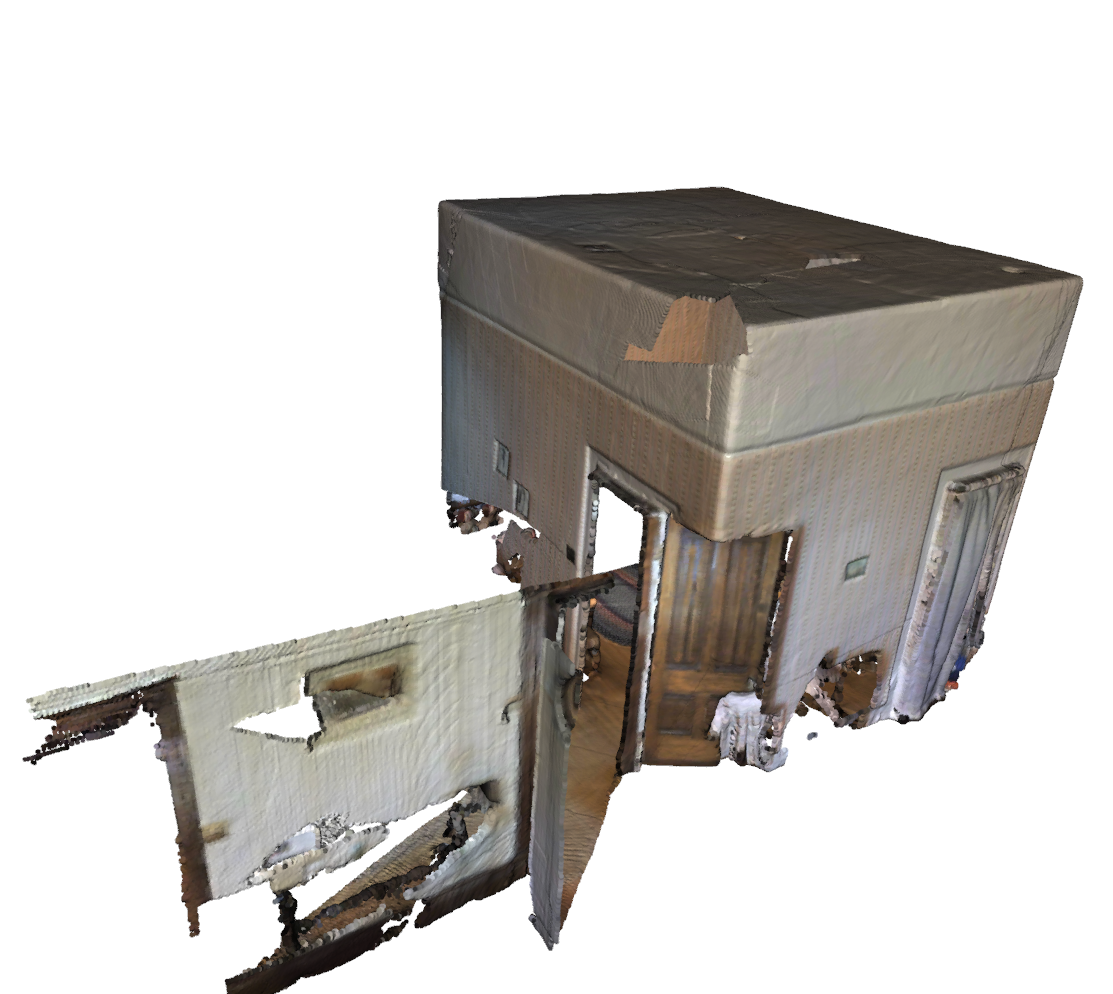}&
	     \includegraphics[width=0.20\textwidth]{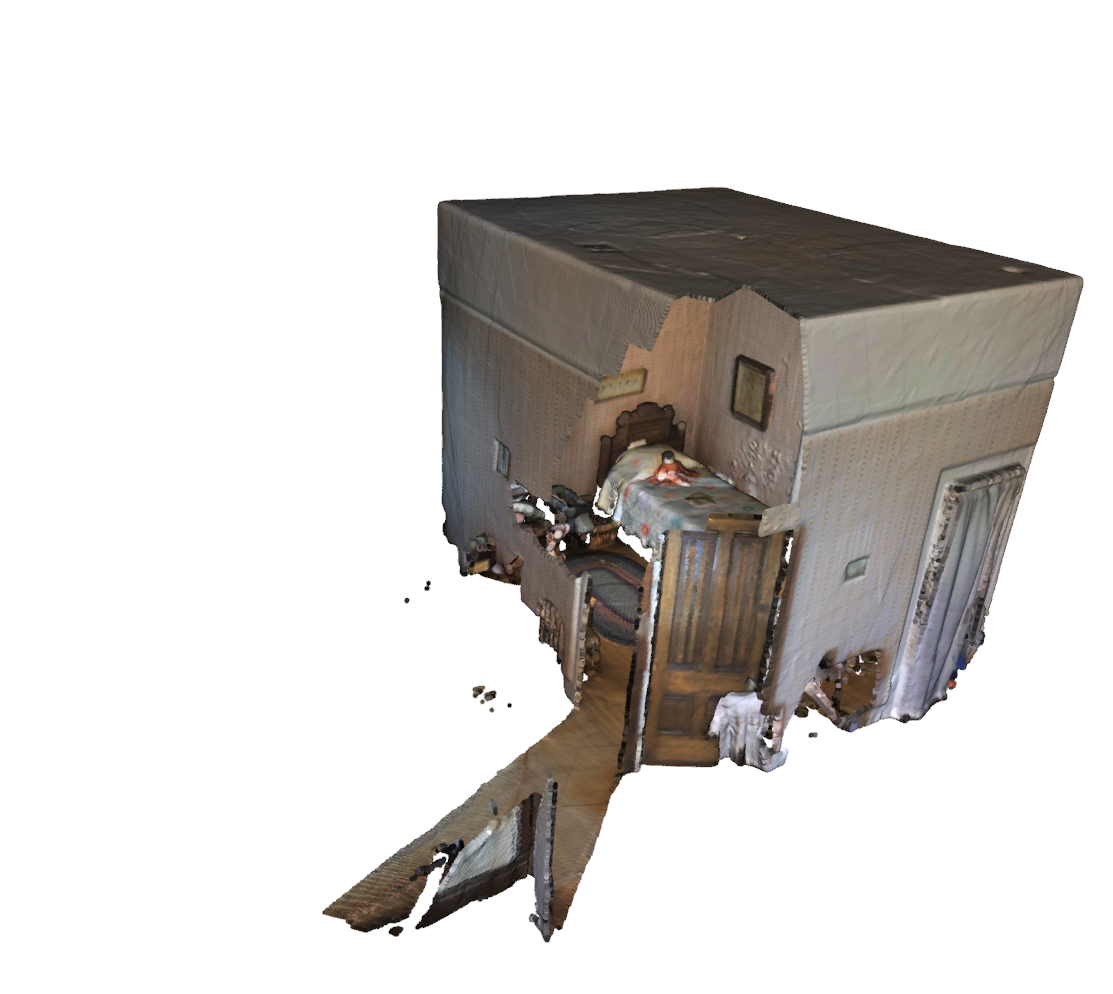}&
		\includegraphics[width=0.20\textwidth]{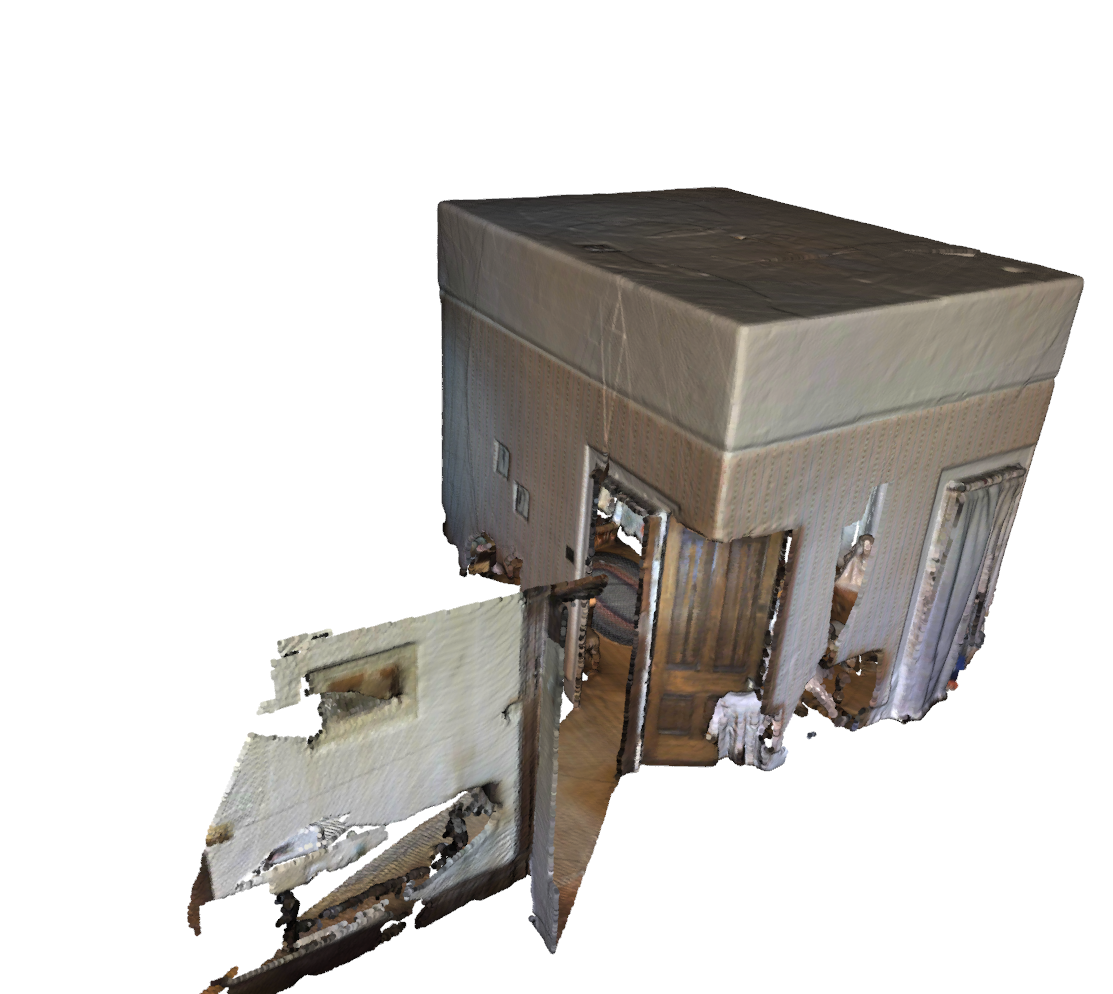}&
		\includegraphics[width=0.20\textwidth]{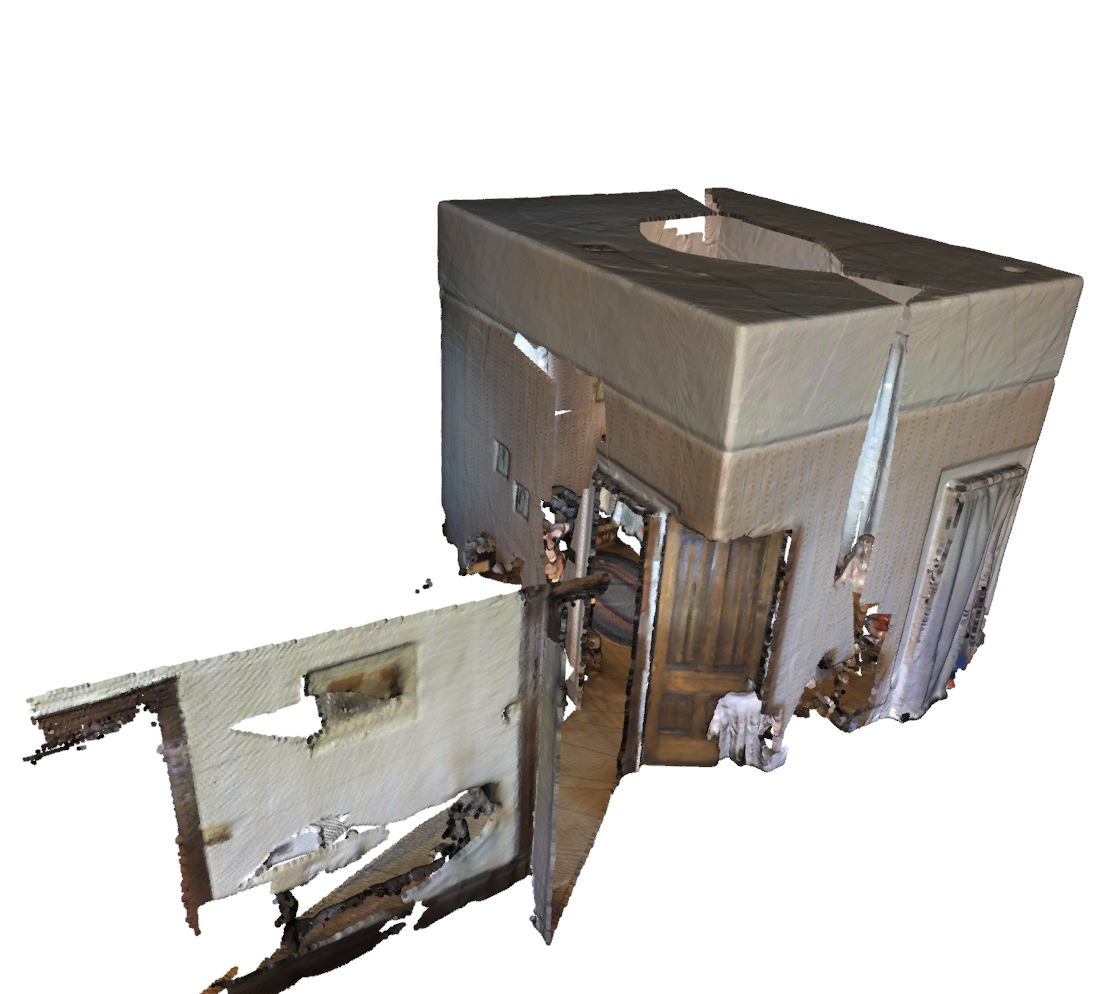}
  	\end{tabular}
\end{center}
\caption{Selected poses on the dome surface and their corresponding 3D reconstruction for a real room in Matterport3D (best viewed in color).}
\label{fig:exp:path}
\end{figure*}
When testing the strategies on real room scans, from Fig.~\ref{fig:exp:coverage_ratio_syn}(b), we can see that our variants CNN2D and CNN5D are still the fastest in the early phase compared to other strategies, although at saturation they are surpassed by BaseGain, Count and CombGain. This suggests that our learning-based strategies may require further domain adaptation when transferring from purely synthetically trained models to real-world scenarios. Fig.~\ref{fig:exp:coverage_ratio_syn}(c,d) show the averaged coverage ratio at saturation and its standard deviation under synthetic and real scenes, respectively. The variance occurs due to different viewpoint initialisation. Fig.~\ref{fig:exp:path} showcases the resulted paths on the dome-surface and their corresponding 3D reconstruction of one real room using various NBV strategies (more results can be found in the supplementary material). In this particular run, we can see that CNN2D is able perform a rather complete exploration of the room.
\begin{table}[!t]
\caption{Processing time for different NBV strategies averaged over all steps tested with the two datasets from SUNCG and Matterport3D (Unit is in second).}
\centering
\resizebox{\textwidth}{!}{
\begin{tabular}{|c|c|c|c|c|c|c|c|c|c|c|}
\hline
  & Random & BaseGain & Count & CombGain & CNNdepth & CNNUtility & CNN2D & CNN5D & MLP2D & MLP5D  \\ \hline
 SUNCG & 0 & 0.028 & 0.016 & 0.040  & 0.007 & 0.018 & 0.023 & 0.024 & 0.058 & 0.098  \\ \hline
 Matterport3D & 0 &  0.022 &0.016 & 0.047 & 0.007 & 0.026 & 0.030 & 0.031 & 0.064 & 0.105 \\ \hline
 \hline
 Average & 0 &  0.025 & 0.016 & 0.044 & 0.007 & 0.022 & 0.027 & 0.028 & 0.061 & 0.102 \\ \hline
\end{tabular}}
\label{table:exp:NBV_time}
\end{table}

\paragraph{\textbf{Computational analysis.}}We performed experiments using a Dell Alienware Aurora with core i7. Table~\ref{table:exp:NBV_time} shows the computational time for different NBV methods, where the processing time includes the time for utility map generation through ray-tracing and the time for NBV estimation. Random uses 0 seconds as the strategy is a random number generator. CNNDepth is fastest ($0.007$s) among all learning-based strategies as it does not require the production of utility map. CNNUtility requires utility map generation therefore is slower than CNNDepth, however is the second fastest as it does not require the preparation of stacking depth and utility map. CNN2D and CNN5D requires almost the same time for the utility map generation and data preparation for passing through the network. CNN2D and CNN5D are slower than Count, comparable to BaseGain and faster than CombGain. CombGain is slow as it performs extra processing on the view-dependent descriptor update for each voxel before generating the utility maps. MLP-based strategies are the slowest as each channel of the input is processed through the pretrained network.

\section{Conclusion}
\label{sec:conclusion}
In this paper we proposed ExHistCNN, a light-weight learning-based solution to address autonomous 3D exploration of any unknown environment. With experiments using dataset from both synthetic and real rooms, we showed that our ExHistCNN, both CNN2D and CNN5D, are able to effectively encode depth observation and reconstruction history for the exploration task. ExHistCNN-based NBV strategies are computationally efficient and able to explore the space faster in the early phase while approaching the oracle NBV performance in the synthetic dataset. When testing with real room scans, our ExHistCNN even if purely trained using synthetic data maintains its property of fast exploration at the early stage, while achieving less final coverage compared to the soa method. As future work, we will further investigate domain adaptation techniques to boost the exploration performance in real-world scenarios.    

\section*{Acknowledgements}
We would like to thank Andrea Zunino, Maya Aghaei, Pietro Morerio, Stuart James and Waqar Ahmed for their valuable suggestions and support for this work. 

\clearpage
\bibliographystyle{splncs04}
\bibliography{6579}
\end{document}